\documentclass{article}

\usepackage{microtype}
\usepackage{graphicx}
\usepackage{subfigure}
\usepackage{booktabs} 
\usepackage{gensymb}

\usepackage{hyperref}

\usepackage[accepted]{icml2019}

\icmltitlerunning{Submission and Formatting Instructions for ICML 2021}

\begin{document}

\twocolumn[
\icmltitle{Precipitaion Nowcasting using Deep Neural Network}
           
\icmlsetsymbol{equal}{*}

\begin{icmlauthorlist}
\icmlauthor{Mohamed Chafik Bakkay}{equal,irt}
\icmlauthor{Mathieu Serrurier}{irit}
\icmlauthor{Valentín Kivachuk Burdá}{irt}
\icmlauthor{Florian Dupuy}{irt}
\icmlauthor{Naty Citlali Cabrera-Gutiérrez}{irt}
\icmlauthor{Michaël Zamo}{meteofrance}
\icmlauthor{Maud-Alix Mader}{irt}
\icmlauthor{Olivier Mestre}{meteofrance}
\icmlauthor{Guillaume Oller}{irt}
\icmlauthor{Jean-Christophe Jouhaud}{cerfacs}
\icmlauthor{Laurent Terray}{cecu}
\end{icmlauthorlist}

\icmlaffiliation{irt}{Institut de Recherche Technologique Saint-Exupéry, Toulouse, France}
\icmlaffiliation{irit}{Institut de Recherche en Informatique de Toulouse, Toulouse, France}
\icmlaffiliation{meteofrance}{Météo-France, Toulouse, France}
\icmlaffiliation{cerfacs}{Centre Européen de Recherche et de Formation Avancée en Calcul Scientifique, Toulouse, France}
\icmlaffiliation{cecu}{Climate, Environment, Coupling and Uncertainties, Toulouse, France}

\icmlcorrespondingauthor{Mohamed Chafik Bakkay}{mohamed-chafik.bakey@irt-saintexupery.com}

\icmlkeywords{Machine Learning, ICML}

\vskip 0.3in
]




\begin{abstract}

Precipitation nowcasting is of great importance for weather forecast users, for activities ranging from outdoor activities and sports competitions to airport traffic management. In contrast to long-term precipitation forecasts which are traditionally obtained from numerical models, precipitation nowcasting needs to be very fast. It is therefore more challenging to obtain because of this time constraint. Recently, many machine learning based methods had been proposed. We propose the use three popular deep learning models (U-net, ConvLSTM and SVG-LP) trained on two-dimensional precipitation maps for precipitation nowcasting. We proposed an algorithm for patch extraction to obtain high resolution precipitation maps. We proposed a loss function to solve the blurry image issue and to reduce the influence of zero value pixels in precipitation maps.

\end{abstract}

\section{Introduction}
\label{sec:intro}

Precipitation nowcasting is the prediction of the future precipitation rate in a given geographical region with an anticipation time of a few hours at most (1-6 hours). With climate change, the frequency of extreme weather events is increasing. It is becoming more and more important to provide usable forecasts at high spatial and temporal resolutions. Such predictions facilitate planning and crisis management. It is of great importance for weather forecast users, for activities ranging from outdoor activities and sports competitions to airport traffic management. 

Long-term precipitation forecasts are traditionally obtained from numerical weather prediction models that explicitly simulate atmospheric physics and can provide reliable predictions. However, these models are very expensive in computation time and take hours to make inference. This limits their ability to be used in precipitation nowcasting which needs to be very fast. Precipitation nowcasting is therefore more challenging to obtain because of time constraint.

Precipitation nowcasting based on radar images is a particular case of spatio-temporal prediction problem which contains two steps. The first step consists on learning the characteristics of a sequence of images in a self-supervised way. In the second step, the learned characteristics are used for the prediction task. In fact, spatio-temporal prediction is an ill-posed problem; there are multiple possible solutions and a machine learning methods aim to find the most probable ones. 

Furthermore, precipitation nowcasting is a more challenging than classical spatio-temporal prediction tasks due to the complexity and the chaotic nature of the atmospheric environment. In fact, clouds have variable speed, and they may accumulate, dissipate or change rapidly~\cite{wang2017predrnn}. Indeed, precipitation maps based on radar images are obtained with a lower frequency (1 frame every 5-15 minutes) than video images (25-30 frames every second). For instance, some computer vision methods are based on optical flow~\cite{bowler2004development} which estimates object movement in a sequence of images. However, optical flow is unable to represent the sudden change in weather since it makes assumptions that are clearly violated, e.g., the amount of rain will not change over time.

Recently, many deep learning based methods had been proposed. These methods are trained on a large amount of data and can learn many patters. For instance, if the neural network has seen sudden changes during training (for example, a sudden agglomeration
of clouds that appears), it can discover these cases and give reasonable predictions.

In this work, we formulate precipitation nowcasting issue as a spatio-temporal prediction problem where both input and prediction target are image sequences. We propose the use three popular deep learning models: a deterministic RNN-based model (ConvLSTM), a stochastic RNN-based model (SVG-LP) and a CNN-based model (U-net). These models are trained on two-dimensional precipitation maps. Unlike Argawal et al.~\yrcite{agrawal2019machine}, we train our models to perform a regression task rather than classification task. Thus, the proposed models generates continuous precipitation values.

Giving high resolution precipitation maps $(1050\times1650)$ as input to the neural networks saturates the GPU memory. One solution to this issue is to resize precipitation maps (using bilinear interpolation for example). The advantage of this solution is that all the available spatial information is given as input to the neural network. However, resizing causes a loss of resolution. Thus, we propose an algorithm for patch extraction. It splits precipitation maps into small patches while including neighboring spatial information in each patch. This allows the network to work with full resolution of the precipitation maps. 

Unlike the mean squared error (MSE) that measures a pixel-to-pixel similarity, structural similarity (SSIM)~\cite{wang2004image} measures the structure similarity between the two images. On the one hand, SSIM loss preserves the structure better than MSE loss. On the other hand, MSE preserves colors and luminance better than SSIM. To capture the best characteristics of both loss functions, we propose to combine them. Indeed, since precipitation maps contains many zero values, using unweighted loss function can erroneous the training. Thus, we added a weighting to the loss that reduces the influence of zero value pixels in precipitation maps.

Our contributions are three folds: (1) We propose the use of three popular deep learning models (U-net, ConvLSTM, SVG-LP) for precipitation nowcasting. (2) We propose an algorithm to extract input patches from precipitation map while including neighboring information for each patch. (3) We propose a loss function to solve the blurry image issue and to reduce the influence of zero value pixels in precipitation maps.

\section{Related Work}
\label{sec:related}

RNN-based methods focus on temporal coherence in a sequence. Deterministic RNN-based methods made simplifying assumption there is only one plausible future. However, this can lead to low-quality predictions in real-world settings with stochastic dynamics. Shi et al.~\yrcite{xingjian2015convolutional} introduced ConvLSTM network for precipitation nowcasting. They added convolution operations in recurring connections of LSTM~\cite{hochreiter1997long}. This model was used by many recent precipitation nowcasting methods~\cite{kumar2020convcast, cao2019video, chen2020deep, gaurprecipitation, yao2020prediction}. Yao et al.~\yrcite{yao2020prediction} added two discriminators to ConvLSTM model in order to ensure continuity of video and realistic of photo. They proposed also an adaptive mean squared error (MSE) loss to reduce the influence of zero value pixels in precipitation maps. Other recent methods such~\cite{adewoyin2020tru, sato2018short} used ConvGRU~\cite{ballas2015delving} which is a variant of the GRU model~\cite{hochreiter1998vanishing} that leverages the convolution operations to encode spatial information. Bonnet et al.~\yrcite{bonnet2020precipitation} used PredRNN++ model~\cite{wang2018predrnn++} which is an improved version of PredRNN model~\cite{wang2017predrnn}. In fact, PredRNN proposed Spatiotemporal LSTM unit (ST-LSTM) that contains double memory mechanism combining time memory updated horizontally with the spatial memory transformed vertically. PredRNN++ makes PredRNN model deeper in time by leveraging a new recurrent structure named Causal LSTM with cascaded dual memories. Some methods~\cite{tran2019computer, franch2020precipitation} used TrajGRU~\cite{shi2017deep} model that can actively learn the location-variant structure for recurrent connections. Tra et al. \yrcite{tran2019computer} included SSIM and multi-scale SSIM to train their models. The predicted radar maps generated by deterministic RNN-based methods are blurry. To handle this problem, stochastic RNN-based methods predict a different possible future for each sample of its latent variables. Bihlo et al.~\cite{bihlo2019precipitation} used SVG-LP~\cite{denton2018stochastic} model for precipitation nowcasting. They used the prior model to learn the basic physical rules that precipitation cells evolve and move in time and space. 

CNN-based methods predict a limited number of images in a single pass. They focus on spatial appearances rather than temporal coherence. They are also used for video prediction, although they only create representations for fixed size inputs. Qui et al.~\yrcite{qiu2017short} use a multi-task CNN that explicitly includes features of the various radar stations to improve their CNN's quality. Ayzel et al.~\yrcite{ayzel2019all} used the all convolutional net~\cite{springenberg2014striving} for precipitation nowcasting. Recently, U-net~\cite{ronneberger2015u} is used by precipitation nowcasting methods\cite{samsi2019distributed, agrawal2019machine, ayzel2020rainnet, trebing2020smaat}. Samsi et al.~\yrcite{samsi2019distributed} proposed a data-parallel model to speed up the training time. Their U-net model takes as input a temporal sequence of $7$ images and outputs $6$ images in the future. They used MSE loss for training. Agrawal et al.~\yrcite{agrawal2019machine} proposed a model that performs three binary classifications that indicate whether the rate exceeds thresholds that roughly correspond to trace rain, light rain and moderate rain. Ayzel et al. proposed RainNet~\cite{ayzel2020rainnet} that predicts continuous radar echo maps at a lead time of five minutes based on the past $4$ consecutive maps. This recursive approach was repeated up to a maximum lead time of $60$ minutes. Unfortunately, the spatial smoothing was an undesirable property of RainNet prediction and became increasingly apparent at longer lead times. Trebing et al.~\yrcite{trebing2020smaat} predict single precipitation map using the past $12$ consecutive maps. They used depthwise-separable convolutions in order to reduce the number of parameters. Some methods~\cite{xiang2020precipitation, tian2020ground} used a CNN models to predict one future precipitation value. Xiang et al.~\yrcite{xiang2020precipitation} proposed a mechanism to deal with missing radar data using the features extracted from the historical radar map sequence. Socaci et al.~\cite{socaci2020xnow} introduced methodology for radar data prediction using the Xception \cite{chollet2017xception} deep learning model.

Recent deep learning models such as graph convolutional networks (GCN)~\cite{kipf2016semi} are used for precipitation nowcasting~\cite{miao2020multimodal}. To handle areas with unavailable precipitation, Miao et al.~\cite{miao2020multimodal} convert all observation information into a graph structure and introduce a semisupervised graph convolutional network with a sequence connect architecture to learn the features of all local areas. In their model, different modalities of observation data (including both meteorological and non meteorological data) are modeled jointly.

\section{Data}

\subsection{Data description}

\begin{figure}[htb]
\centering
\begin{minipage}[b]{0.49\linewidth}
  \centering
  \centerline{\includegraphics[width=\linewidth]{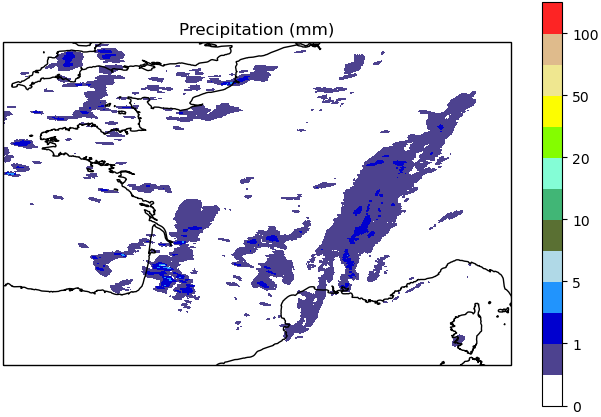}}
  \centerline{\footnotesize (a) Precipitation}
\end{minipage}
\begin{minipage}[b]{0.49\linewidth}
  \centering
  \centerline{\includegraphics[width=\linewidth]{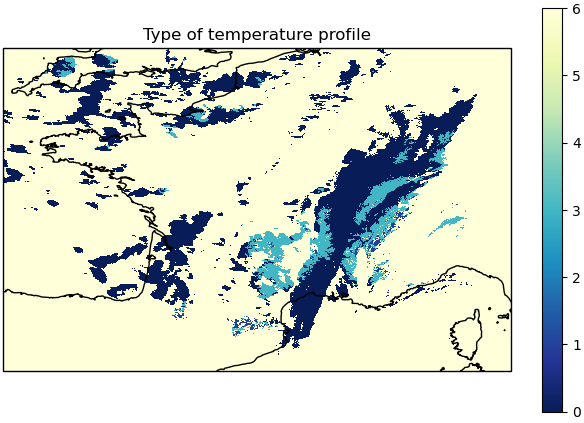}}
  \centerline{\footnotesize (b) Type of temperature profile}
\end{minipage}
\begin{minipage}[b]{0.49\linewidth}
  \centering
  \centerline{\includegraphics[width=\linewidth]{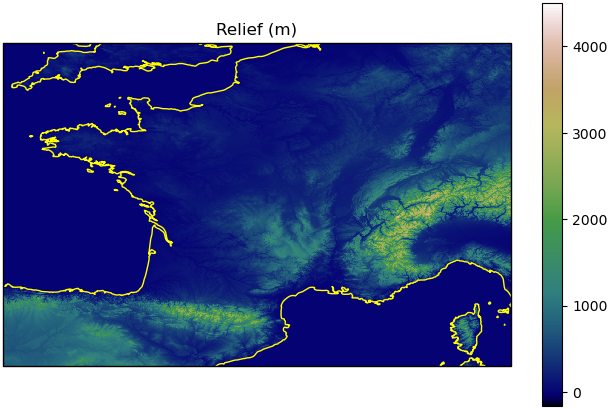}}
  \centerline{\footnotesize (c) Relief}
\end{minipage}
\caption{Input data.}
\label{fig:variables}
\end{figure}

The dataset used in this paper is based on radar echo. It is collected by METEO FRANCE in France from 2017 to 2018. This dataset contains two types of data: ANTILOPE and HYDRE. ANTILOPE includes convective precipitation, precipitation and the duration of convective precipitation. ANTILOPE data is recorded every $15$ minutes. HYDRE includes temperature profile type and precipitation type. HYDRE data is recorded every 5 minutes. METEO FRANCE dataset covers an area of around $1000\times1000 km^2$ (latitude from $41.00\degree N$ to $51.50\degree N$ and longitude from $6.00\degree E$ to $10.50\degree W$) centered in France. Since the spatial resolution is $0.01\degree$, the map size is $1050\times1650$. For relief, we used GMTED2010 dataset~\cite{danielson2011global} which have a resolution of 30 $arc-sec$ ($\approx0.00833\degree$). We interpolate relief maps to obtain a resolution of $0.01\degree$. We used three input data: precipitation (Figure~\ref{fig:variables}.(a)), type profile of temperature (Figure~\ref{fig:variables}.(b)) and relief (Figure~\ref{fig:variables}.(c)).

\subsection{Data partitioning}

\begin{figure}[htb]
\centering\includegraphics[width=\linewidth]{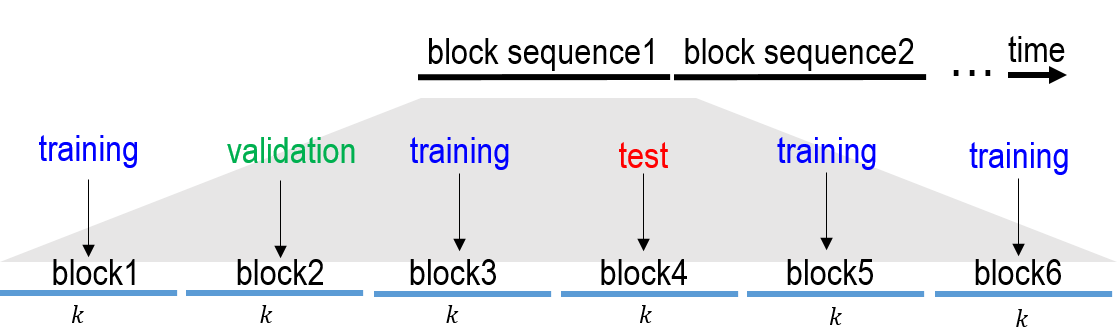}
\caption{Example of data partitioning.}
\label{fig:partition}
\end{figure}

We select seven months (January-July 2018) from our dataset. Since data is recorded every $15$ minutes, there are $96$ images per day. We obtain a total of $20352$ images. To get disjoint subsets for training, testing and validation, we partition the data into block sequences (Figure~\ref{fig:partition}). Each block sequence contains $6$ blocks of $k=47$ images. For each block sequence, we randomly assign $4$ blocks for training, $1$ block for testing and $1$ block for validation. The image sequences used in the neural network are sliced from these blocks using a sliding window of size $s = s_{in} + s_{out} = 12$ ($s_{in}$ is the size input sequence and $s_{out}$ is the size of output sequence). 

\subsection{Data preprocessing}

Giving high resolution precipitation maps $(1050\times1650)$ as input to the neural networks saturates the GPU memory. 

\subsubsection{Full map resize}

The first solution to this issue is to resize maps to ($256\times256$) using bilinear interpolation (Figure~\ref{fig:data_processing}.(a)). The advantage of this method is that all available spatial information is given to the neural network. The downside is that the interpolation causes a loss of resolution.

\subsubsection{Patch extraction}

\begin{figure}[htb]
\centering
\begin{minipage}[b]{\linewidth}
  \centering
  \centerline{\includegraphics[width=\linewidth]{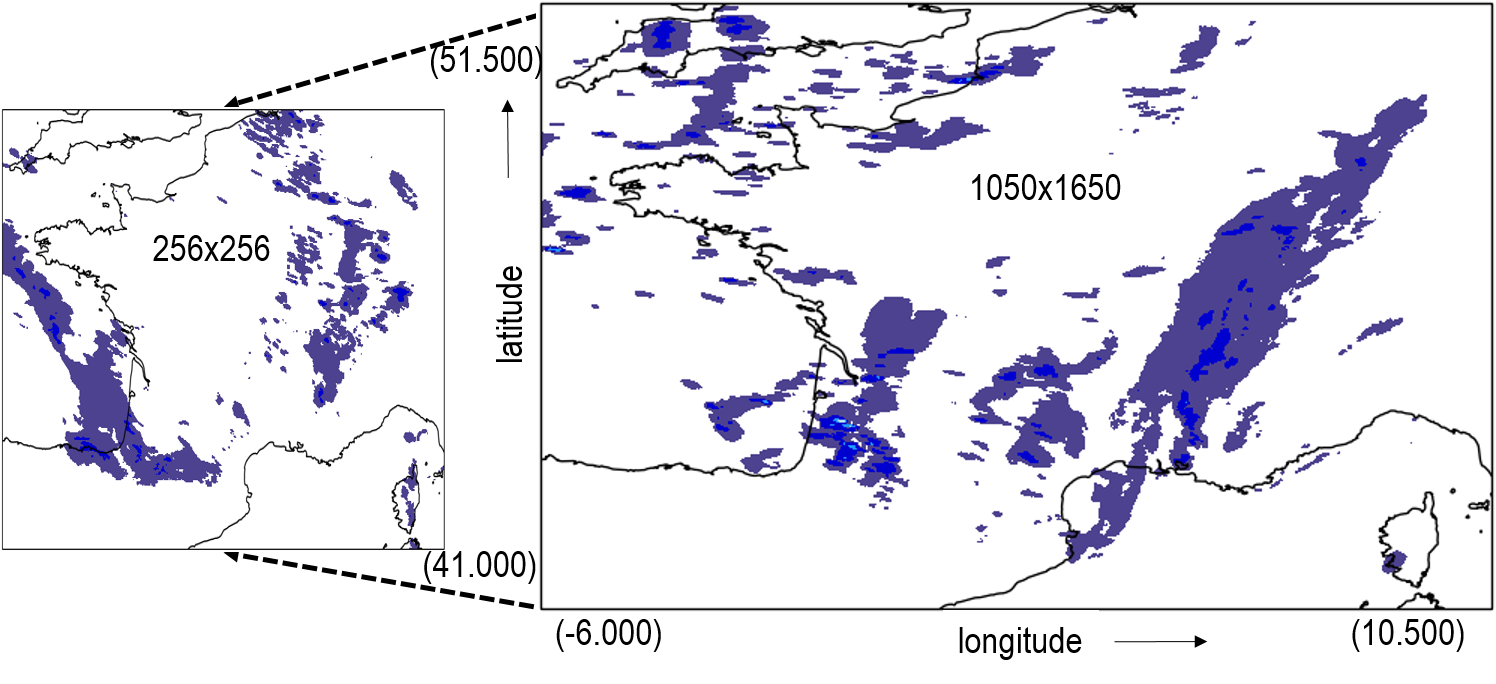}}
  \centerline{\footnotesize (a) Full map resize}
\end{minipage}
\begin{minipage}[b]{\linewidth}
  \centering
  \centerline{\includegraphics[width=\linewidth]{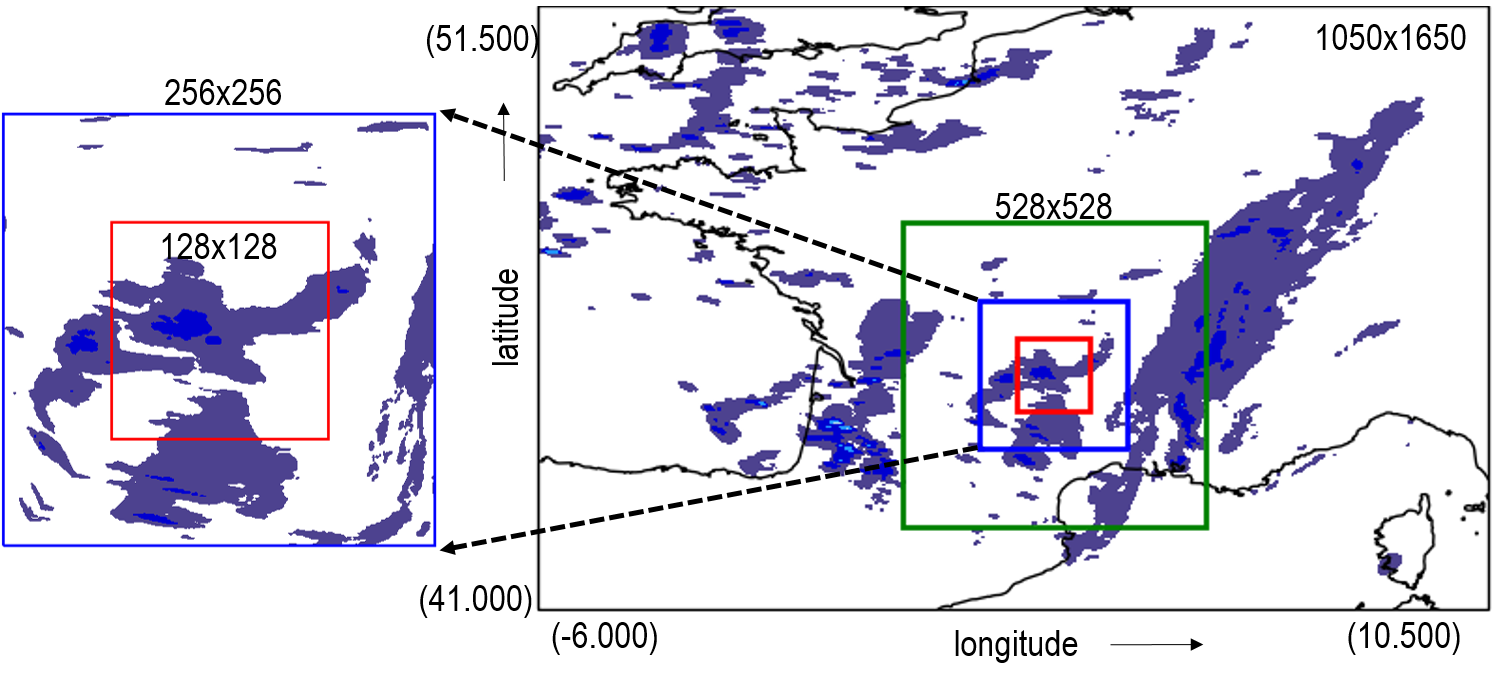}}
  \centerline{\footnotesize (b) Patch extraction}
\end{minipage}
\caption{Data preprocessing}
\label{fig:data_processing}
\end{figure}

\begin{figure}[htb]
  \centering
  \centerline{\includegraphics[width=\linewidth]{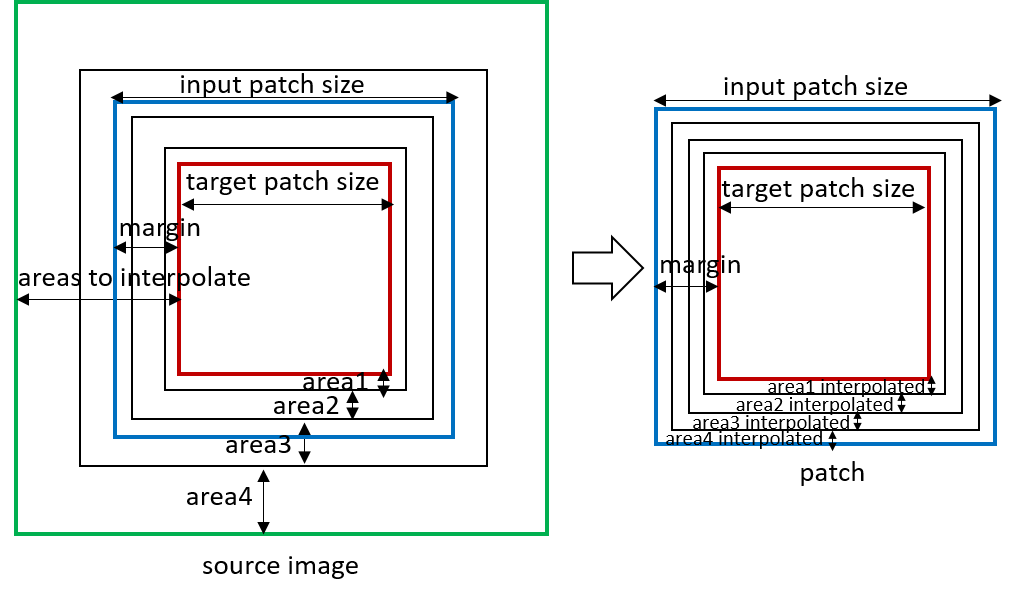}}
  \caption{Patch extraction} 
  \label{fig:patch_extaction}
\end{figure}

The second solution consists on partitioning the map into ($256\times256$) patches and making predictions for each patch. However, these predictions are dependent. In fact, since clouds moves over time, the prediction of precipitation inside each patch depends on information in neighboring patches. Thus, neighborhood information should be included in each input patch and should be used for prediction. We propose an algorithm (Algorithm~\ref{alg:patch_extraction}) to extract patches. It consists in interpolating the neighborhood information (the areas between the green rectangle and the red rectangle) and including this information inside the input patch (the area between the blue rectangle and the red rectangle) (Figure~\ref{fig:data_processing}.(b)). Each area is interpolated independently. Since the near areas are more important for the prediction, the further away from the target patch (red rectangle), the more the size of the area to interpolate is increased (Figure~\ref{fig:patch_extaction}).

\begin{algorithm}
   \caption{Patch extraction}
   \label{alg:patch_extraction}
\begin{algorithmic}
   \STATE {\bfseries Input:} source image $I$, patch location in the source image $(x,y)$, input patch size $iSize$, target patch size $tSize$, increment step $step$, increment frequency $freq$

   \STATE $P=empty(iSize, iSize)$ \hfill\small\texttt{//patch}
   \STATE $margin = (iSize-tSize)/2$ \hfill\small\texttt{//margin width}
   \STATE $w_I= 1$ \hfill\small\texttt{//area width}
   \STATE $w_P = 1$ \hfill\small\texttt{//interpolated area width}
   \STATE \small\texttt{//source image indexes}
   \STATE $UR_{I} = x+margin$  \hfill\small\texttt{//upper row}
   \STATE $DR_{I} = x+margin+tSize$ \hfill\small\texttt{//lower row}
   \STATE $LC_{I} = y+margin$ \hfill\small\texttt{//left column}
   \STATE $RC_{I} = y+margin+tSize$ \hfill\small\texttt{//right column}
   \STATE \small\texttt{//patch indexes}
   \STATE $UR_{P} = margin$  \hfill\small\texttt{//upper row}
   \STATE $DR_{P} = margin+tSize$ \hfill\small\texttt{//lower row}
   \STATE $LC_{P} = margin$ \hfill\small\texttt{//left column}
   \STATE $RC_{P} = margin+tSize$ \hfill\small\texttt{//right column}
  \STATE \small\texttt{//copy the target patch} 
  \STATE $P[UR_{P}:DR_{P}, LC_{P}:RC_{P}] = I[ UR_{I}:DR_{I}, LC_{I}:RC_{I}]$
  \STATE \small\texttt{//interpolate each area of the source image}
 \FOR{$k=1$ {\bfseries to} $margin$}
  \STATE \small\texttt{//the area to interpolate is divided into 4 rectangles: upper, lower, left and right}
  \STATE \small\texttt{//interpolate upper rectangle}
  \STATE $Rec = I[UR_{I}-w_{I}:UR_{I}, LC_{I}-w_{I}:RC_{I}+w_{I}]$ 
  \STATE $Rec = interpolate(Rec,(w_{P}, RC_{P}-LC_{P}+2*w_{P}))$
  \STATE $P[UR_{P}-w_{P}:UR_{P},LC_{P}-w_{P}:RC_{P}+w_{P}] = Rec$
  
  \STATE \small\texttt{//interpolate lower rectangle}
  
  \STATE $Rec = I[DR_{I}:DR_{I}+w_{I}, LC_{I}-w_{I}:RC_{I}+w_{I}]$ 
  \STATE $Rec = interpolate(Rec,(w_{P}, RC_{P}-LC_{P}+2*w_{P}))$
  \STATE $P[DR_{P}:DR_{P}+w_{P},LC_{P}-w_{P}:RC_{P}+w_{P}] = Rec$
  
  \STATE \small\texttt{//interpolate left rectangle}
  
  \STATE $Rec = I[UR_{I}-w_{I}:DR_s+w_{I}, LC_{I}-w_{I}:LC_{I}]$ 
  \STATE $Rec = interpolate(Rec,(DR_{P}-UR_{P}+2*w_{P}, w_{P}))$
  \STATE $P[UR_{P}-w_{P}:DR_{P}+w_{P},LC_{P}-w_{P}:LC_{P}] = Rec$ 
  
  \STATE \small\texttt{//interpolate right rectangle}
  
  \STATE $Rec = I[UR_{I}-w_{I}:DR_{I}+w_{I}, RC_{I}:RC_{I}+w_{I}]$ 
  \STATE $Rec = interpolate(Rec,(DR_{P}-UR_{P}+2*w_{P}, w_{P}))$
  \STATE $P[UR_{P}-w_{P}:DR_{P}+w_{P},RC_{P}:RC_{P}+w_{P}] = Rec$ 
  \STATE\small\texttt{//increment area width}
  \IF{$k\%freq==0$}
  \STATE $w_I = w_I + step$ 
  \ENDIF
   \STATE \small\texttt{//increment source image indexes}
   
   \STATE $UR_{I} = UR_{I} - w_{I}$
   \STATE $DR_{I} = DR_{I} + w_{I}$
   \STATE $LC_{I} = LC_{I} - w_{I}$
   \STATE $RC_{I} = RC_{I} + w_{I}$
   
   \STATE \small\texttt{//increment patch indexes}
   
   \STATE $UR_{P} = UR_{P} - w_{P}$
   \STATE $DR_{P} = DR_{P} + w_{P}$
   \STATE $LC_{P} = LC_{P} - w_{P}$
   \STATE $RC_{P} = RC_{P} + w_{P}$
   
 \ENDFOR
\end{algorithmic}
\end{algorithm}

\subsection{Data transformation and normalization}

\begin{figure}[htb]
  \centering
  \begin{minipage}[b]{0.49\linewidth}
  \centering
  \centerline{\includegraphics[width=\linewidth]{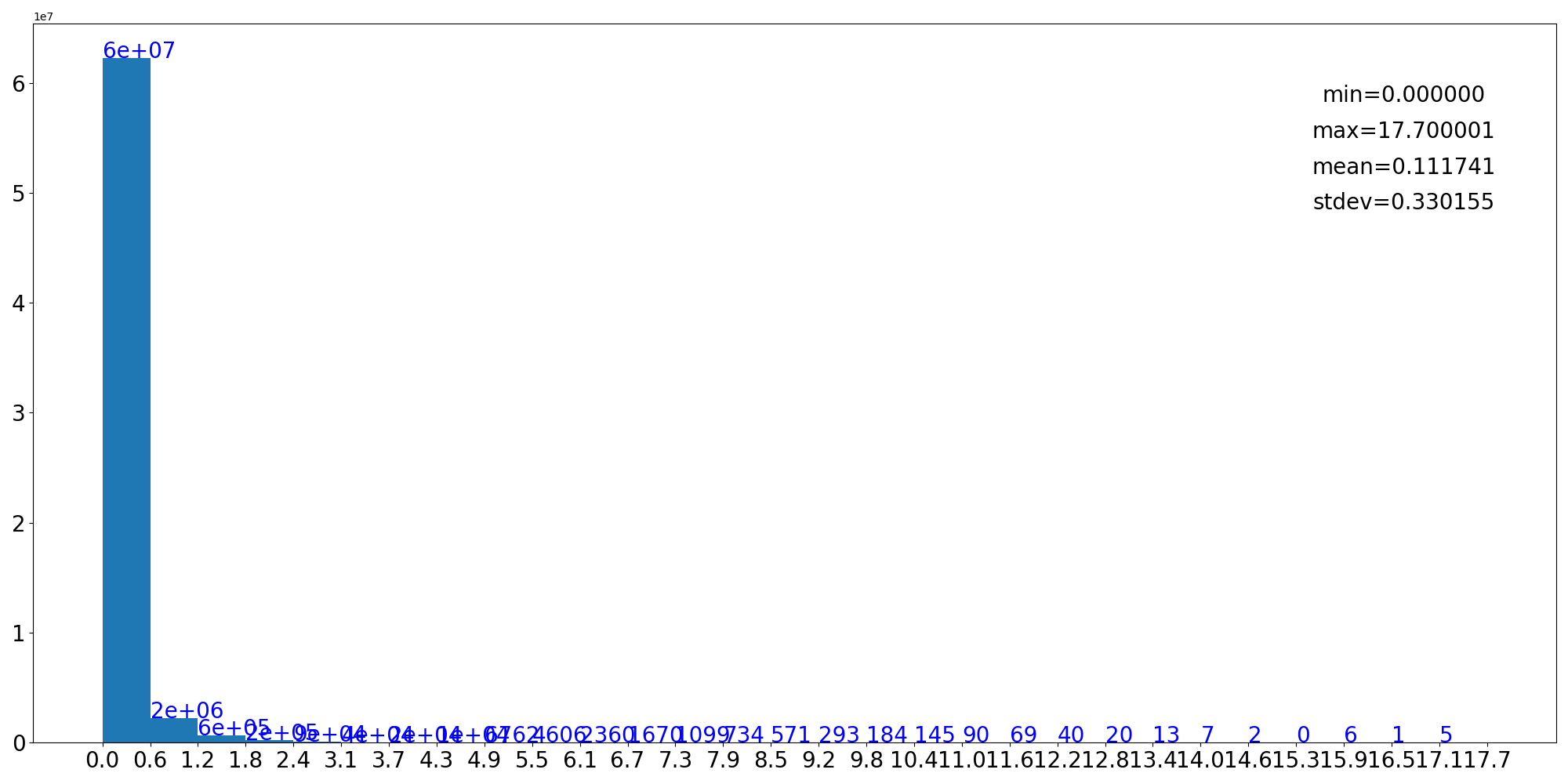}}
  \centerline{\footnotesize (a) original distribution}
  \end{minipage}
  \begin{minipage}[b]{0.49\linewidth}
  \centering
  \centerline{\includegraphics[width=\linewidth]{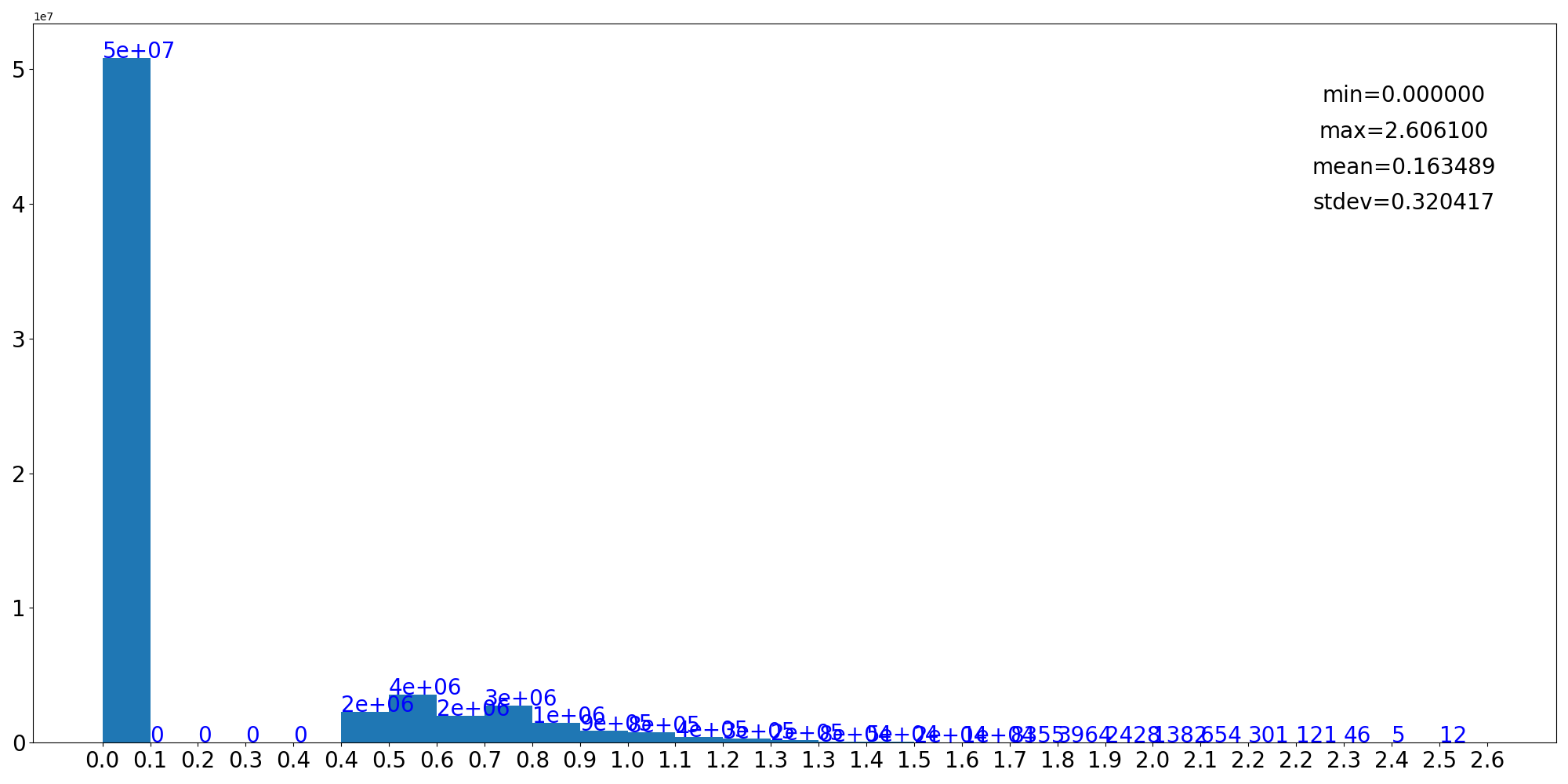}}
  \centerline{\footnotesize (b) root cube}
  \end{minipage}
  \begin{minipage}[b]{0.49\linewidth}
  \centering
  \centerline{\includegraphics[width=\linewidth]{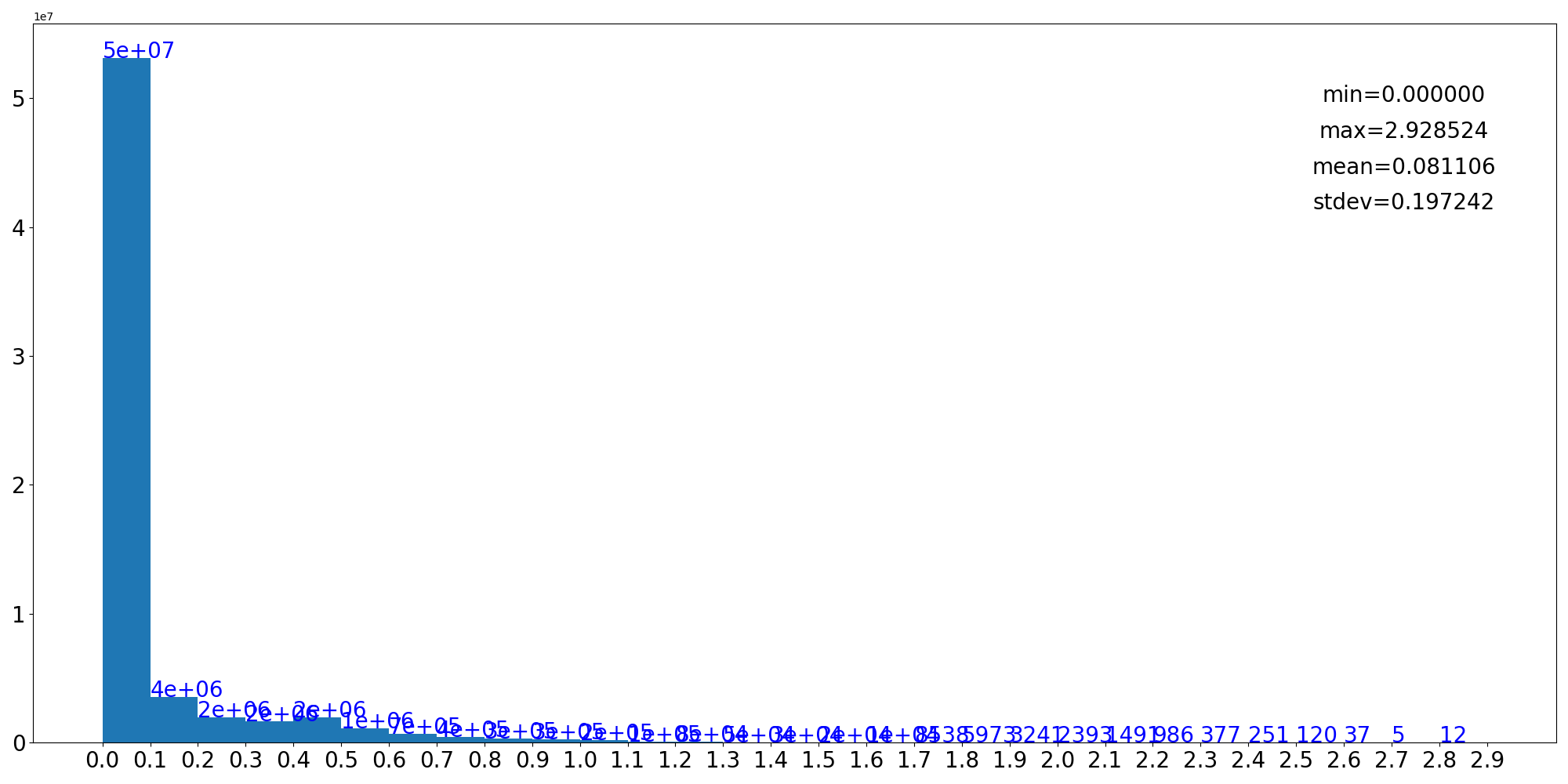}}
  \centerline{\footnotesize (c) log(x+1)}
  \end{minipage}
    \begin{minipage}[b]{0.49\linewidth}
  \centering
  \centerline{\includegraphics[width=\linewidth]{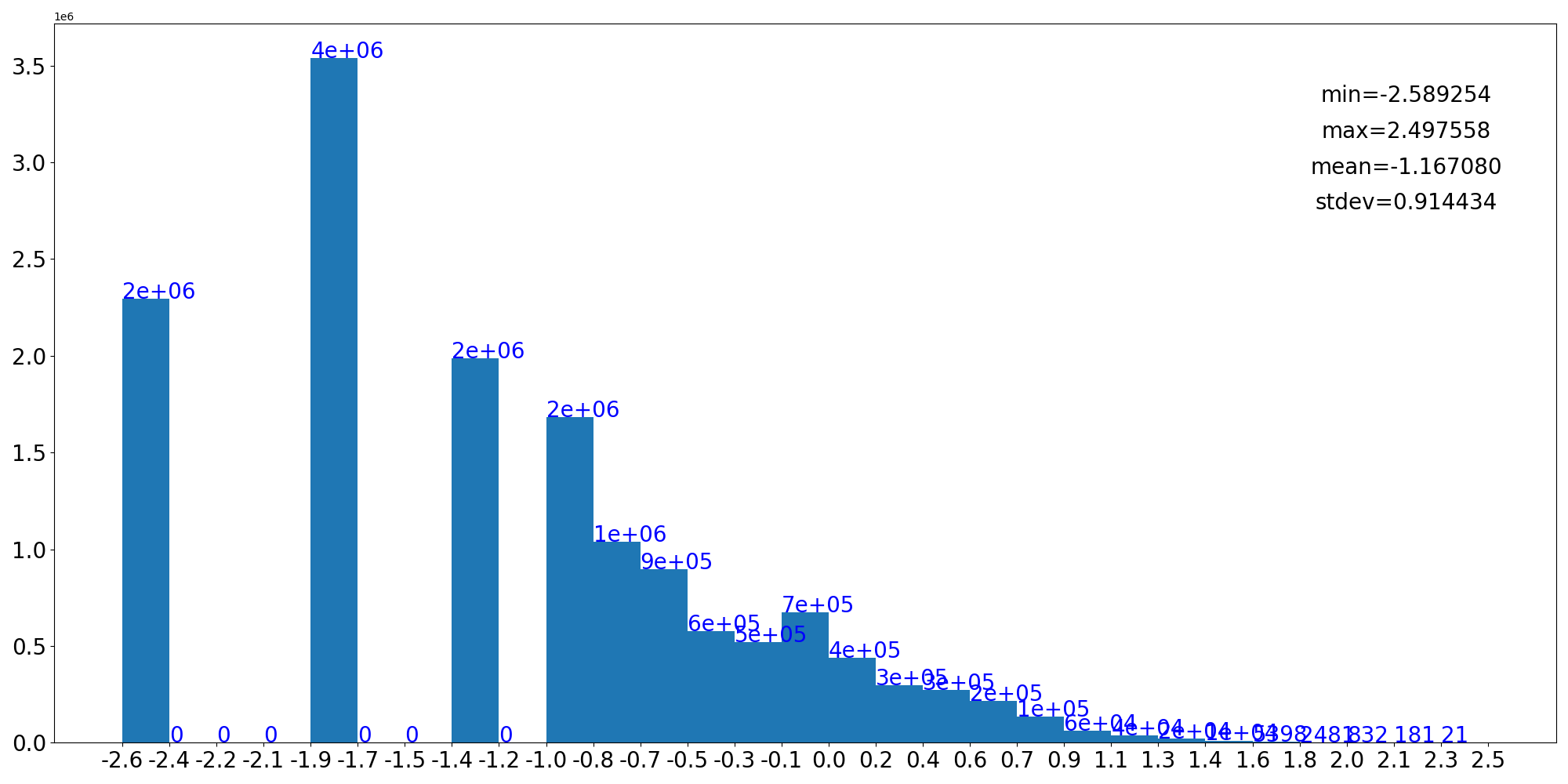}}
  \centerline{\footnotesize (d) box cox}
  \end{minipage}
  \caption{Distribution of precipitation for two months} 
  \label{fig:transformation}
\end{figure}

Considerable right skewness in precipitation distributions arises because of the relatively numerous occurrences of ``no-rain'' days \cite{skees1974comments}. In fact, most precipitation values fall between $0$ and $1$, see Fig.~\ref{fig:transformation}.(a). In general, right skewness should be reduced to make patterns in the data more interpretable and helps to meet the assumptions of inferential statistics \cite{gokmen2015acrylamide}. For that, many transformations can be used such as log transformation, square root transformation and Box Cox\cite{box1964analysis}. In our case, the ``spike'' of zeros all have the same value. Thus, no matter what transformation we apply, they are always going to have the same value. They are always going to be a ``spike'' sticking out from the rest of the distribution. Thus, we choose to apply log transformation to compress high values of precipitation~\cite{mccune2002analysis} and then to avoid that these values will be considered as noise by the neural network. In order to include zero values, we used $log(x+1)$ instead of $log(x)$. Finally, we standardize all variables to have a mean of zero and a standard deviation of $1$ using z-score:
\begin{equation}
   var_{standardized}=\frac{var-mean(var)}{stdev(var)},
    \label{equation:z-score}
\end{equation}

\section{Models}

\begin{figure}[htb]
\begin{minipage}[b]{\linewidth}
  \centering
  \centerline{\includegraphics[width=\linewidth]{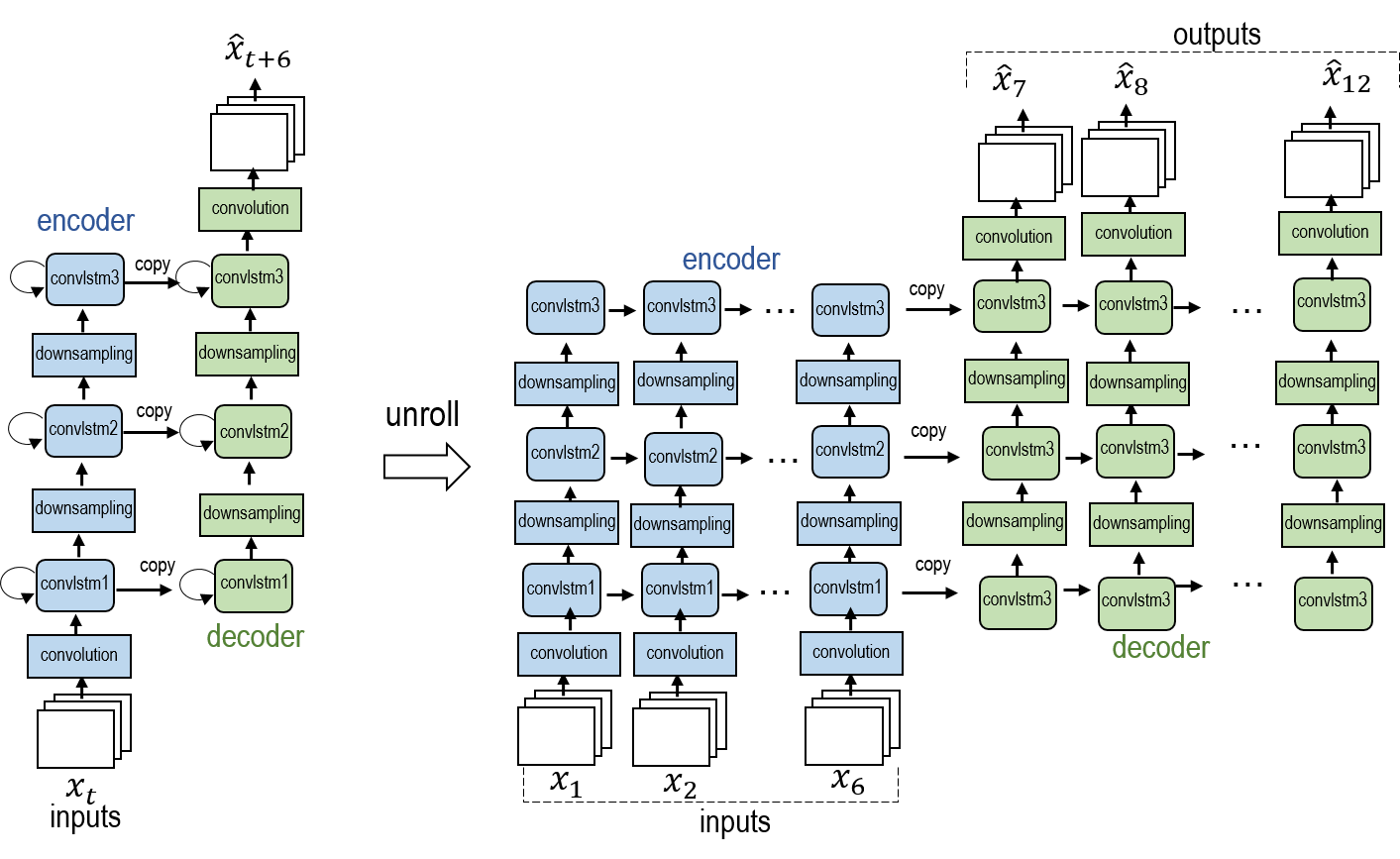}}
  \centerline{\footnotesize (a) deterministic RNN: ConvLSTM}
\end{minipage}
\begin{minipage}[b]{\linewidth}
  \centering
  \centerline{\includegraphics[width=\linewidth]{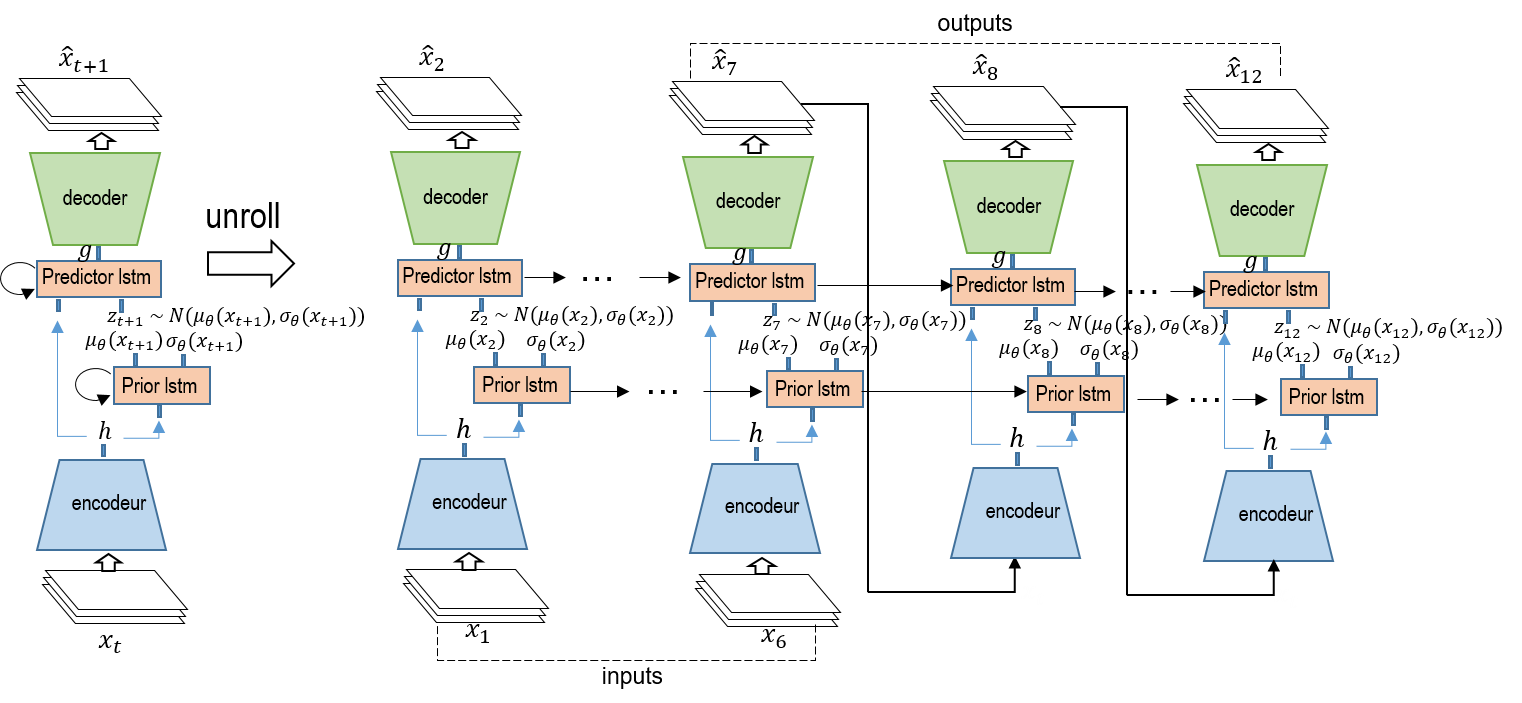}}
  \centerline{\footnotesize (b) stochastic RNN: SVG-LP}
\end{minipage}
\begin{minipage}[b]{\linewidth}
  \centering
  \centerline{\includegraphics[width=\linewidth]{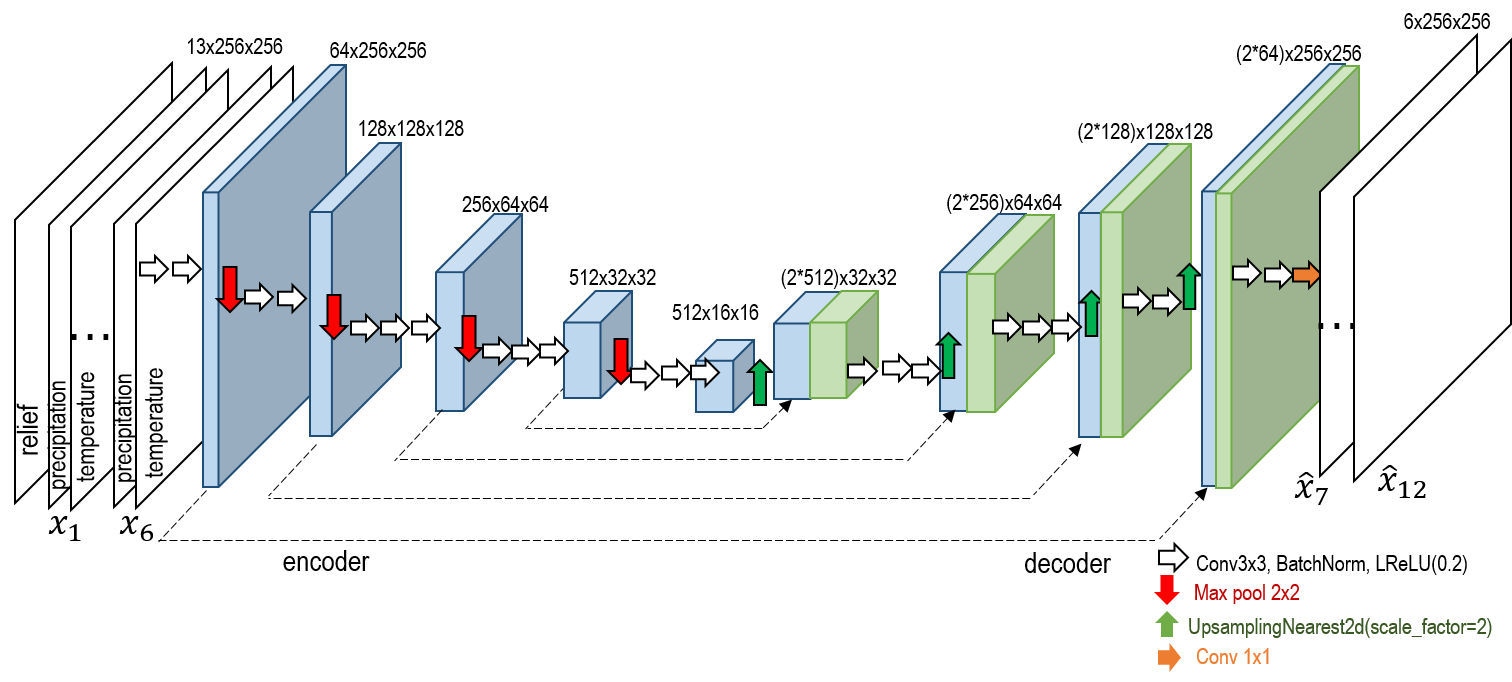}}
  \centerline{\footnotesize (c) CNN: U-net}
\end{minipage}
\caption{Used models.}
\label{fig:models}
\end{figure}

The deterministic RNN-based model (Figure~\ref{fig:models}.(a)) is ConvLSTM \cite{xingjian2015convolutional}. It follows ``sequence to sequence'' architecture. It consists of two networks, an encoding network and a decoding network. the initial states and cell outputs of the decoding network are copied from the last state of the encoding network. Both networks are formed by stacking several ConvLSTM layers. The encoder network consists of three stacked stride-1 convLSTM layers with 64, 192 and 192 filters, respectively. The filter size is $3\times3$ for the three layers. We added a downsampling layer before each convLSTM layer. Each downsampling layer contains a stride-2 $4\times4$ convolution followed by Leaky-ReLU non-linearity. The decoder network consists of three stacked stride-1 convLSTM layers with 192, 192 and 64 filters, respectively. The filter size is $3\times3$ for the three layers. We added a upsampling layer after each convLSTM layer. Each downsampling layers contain a stride-2 $4\times4$ deconvolution followed by Leaky-ReLU non-linearity. After the last dwonsampling layer, we added a stride-1 $3\times3$ convolution followed by Leaky-ReLU non-linearity and a stride-1 $1\times1$ convolution $1\times1$ convolutional layer to generate the final prediction.

The stochastic RNN-based model (Figure~\ref{fig:models}.(b)) is SVG-LP \cite{denton2018stochastic}. The predictor network is a two layer LSTMs with $256$ cells in each layer. The posterior and prior networks are both single layer LSTMs with $256$ cells in each layer. Each network has a linear embedding layer and a fully connected output layer. The output dimensionalities of the LSTM networks are $|g|= 512$; $|\mu_{posterior}| = |\mu_{prior}| = 256$. The encoder and decoder has the same architecture of the proposed U-net. We replaced the last layer of encoder by a stride-1 $16\times16$ convolution followed by batch normalization and tanh nonlinearity. The encoder outputs a vector with dimensionality $|h| = 512$. We added a $16\times16$ deconvolution at the beginning of the decoder to output $512$ feature maps with a $16\times16$ size. This model follows ``sequence modeling'' architecture also named ``one to many'' architecture used in natural language processing (NLP). To generate a sequence of $6$ images (the future), we give the first $6$ images (the past) as input to the network. Starting from $6$ time step, the generated output is passed as input into the next time step.

The CNN-based model (Figure~\ref{fig:models}.(c)) follows an encoder-decoder architecture of U-net network with skip connections~\cite{ronneberger2015u}. In this work, the encoder includes five convolutional layers. The first four convolutional layers are each followed by a downsampling layer that decrease the size of the output feature maps. The first two convolutional layer use two $3\times3$ convolutions. The remaining convolutional layers use three $3\times3$ convolutions. The encoder outputs $512$ feature maps with a $16\times16$ size. The decoder uses four upsampling layers each followed by a convolutional layer to construct an output image with the same resolution of the input one. The first two convolutional layers of decoder use three $3\times3$ convolutions. The third convolutional layers of decoder use two $3\times3$ convolutions. The last convolutional layer of decoder uses three $3\times3$ convolutions to generate $64$ feature maps with a $256\times256$ size then uses a $1\times1$ convolution to produce the finals $6$ precipitation maps. The weights of the network are randomly initialized. Each $3\times3$ convolution is followed by batch normalization and Leaky-ReLU non-linearity. The final $1\times1$ convolution is followed by linear activation.

\section{Loss function}

The network have to learn to produce images having the same structure as ground truth images. This goal can be achieved by using loss function that preserves the structure as is the case with SSIM loss. SSIM score \cite{wang2004image} is used to measure the similarity of structure between the two images, rather than a pixel-to-pixel difference like the PSNR does for example. It is computed between two windows $x_j$ and $y_j$ in the image. It is given by (\ref{equation:ssim}) which combines three comparative measures: luminance (\ref{equation:luminance}), contrast (\ref{equation:contrast}) and structure (\ref{equation:structure}).
\begin{equation}
    SSIM(x_j,y_j)=[l(x_j,y_j)]^\alpha\cdot[c(x_j,y_j)]^\beta\cdot[s(x_j,y_j)]^\gamma, \label{equation:ssim}
\end{equation}
where they set $\alpha=\beta=\gamma=1$
\begin{equation}
    l(x_j,y_j)=\frac{2\mu_{x_j}\mu_{y_j}+C_1}{\mu_{x_j}^2+\mu_{y_j}^2+C_1}, \label{equation:luminance}
\end{equation}
where $C_1=(k_1L)^2$ to stabilize the division with weak denominator, $L$ is the dynamic range of the pixel-values, $k_1=0.01$. 
\begin{equation}
    c(x_j,y_j)=\frac{2\sigma_{x_j}\sigma_{y_j}+C_2}{\sigma_{x_j}^2+\sigma_{y_j}^2+C_2}, \label{equation:contrast}
\end{equation}
where $C_2=(k_2L)^2$ to stabilize the division with weak denominator, $L$ is the dynamic range of the pixel-values, $k_2=0.03$. 
\begin{equation}
    s(x_j,y_j)=\frac{2\sigma_{x_jy_j}+C_3}{\sigma_{x_j}\sigma_{y_j}+C_3}, \label{equation:structure}
\end{equation}
where $C_3=C_2/2$ used to stabilize the division with weak denominator.

In SSIM paper \cite{wang2004image}, in order to evaluate the overall image quality, the mean SSIM score, an uniformly weighted average of the different window scores is defined as:
\begin{equation}
    SSIM(X,Y)=\frac{1}{M} \sum_{j=1}^MSSIM(x_j,y_j), \label{equation:dc}
\end{equation}
where $X$ and $Y$ are the reference and the distorted images, respectively; and $x_j$ and $y_j$ are the image contents at the jth local window; and $M$ is the number of local windows of the image. However, there is high class imbalance: about $98\%$ of the pixels have no precipitation. Thus, with an uniformly weighted loss function, each window contributes equally to the loss function, and a network can achieve high accuracy by simply predicting the dominant class for all pixels. To improve upon this situation, we use a weighted loss calculation in which the loss for each window is weighted based on its standard deviation. We compute a weighted average of the different windows in the image:
\begin{equation}
   WSSIM(X,Y) = \sum_{j=1}^Mw_j*SSIM(x_j,y_j),
    \label{equation:wssim}
\end{equation}
where $w_j=\frac{1+\sigma_{x_j}}{\sum_i^M{1+\sigma_{x_i}}}$, $\sigma_{x_j}$ is the standard deviation of the window $x_j$. For each target window $x_j$, the greater its standard deviation $\sigma_{x_j}$, the greater the variation between its pixels, the greater its weight $w_j$. If $x_j$ does not contain precipitation, then $\sigma_{x_j}=0$ and $w_j=1$. The weighted SSIM loss is defined as :
\begin{equation}
   \ell_{WSSIM}= 1-WSSIM,
    \label{equation:dc}
\end{equation}

Similarly, we propose a weighted version of the MSE loss that reduces the influence of zero value pixels in precipitation maps:
\begin{equation}
   \ell_{WMSE}(X,Y) = \sum_{i=1}^Nw_i*MSE(p_i,q_i),
    \label{equation:wssim}
\end{equation}
where $X$ and $Y$ are the reference and the generated images respectively; and $p_i$ and $q_i$ are their ith pixels respectively; and $N$ is the number of pixel in each image, $w_i=1$ if $p_i < T$, $w_i=3$ otherwise; $T=0.1$ is the ``rain/no-rain'' threshold.

On the one hand, the weighted SSIM loss $\ell_{WSSIM}$ preserves the structure and contrast in high-frequency regions better than the weighted MSE loss $\ell_{WMSE}$. On the other hand, $\ell_{WMSE}$ preserves colors and luminance better than $\ell_{WSSIM}$. To capture the best characteristics of both error functions, we propose to combine them:
\begin{equation}
   \ell_{total}(X,Y)= \alpha \cdot \ell_{WSSIM}(X,Y) + (1-\alpha) \cdot \ell_{WMSE}(X,Y),
    \label{equation:dc}
\end{equation}

We empirically set $\alpha=0.84$. Finally, we have added the regularization term to the loss. Regularization term penalize all the weights of the network $\theta_i$ by making them small making the model simpler and avoiding overfitting.
\begin{equation}
   \ell_{total}= \alpha \cdot \ell_{WSSIM} + (1-\alpha) \cdot \ell_{WMSE} + \beta \cdot \sum_{i=1}^n{\theta_i}^2,
    \label{equation:dc}
\end{equation}
where $\beta = 10^{-3}$ is the penalty term for regularization parameter which determines how much to penalizes the weights; $n$ in the number of the weights in the network. 

\section{Experiments}

\begin{figure}[htb]
\centering\includegraphics[width=8.5cm, height=5cm]{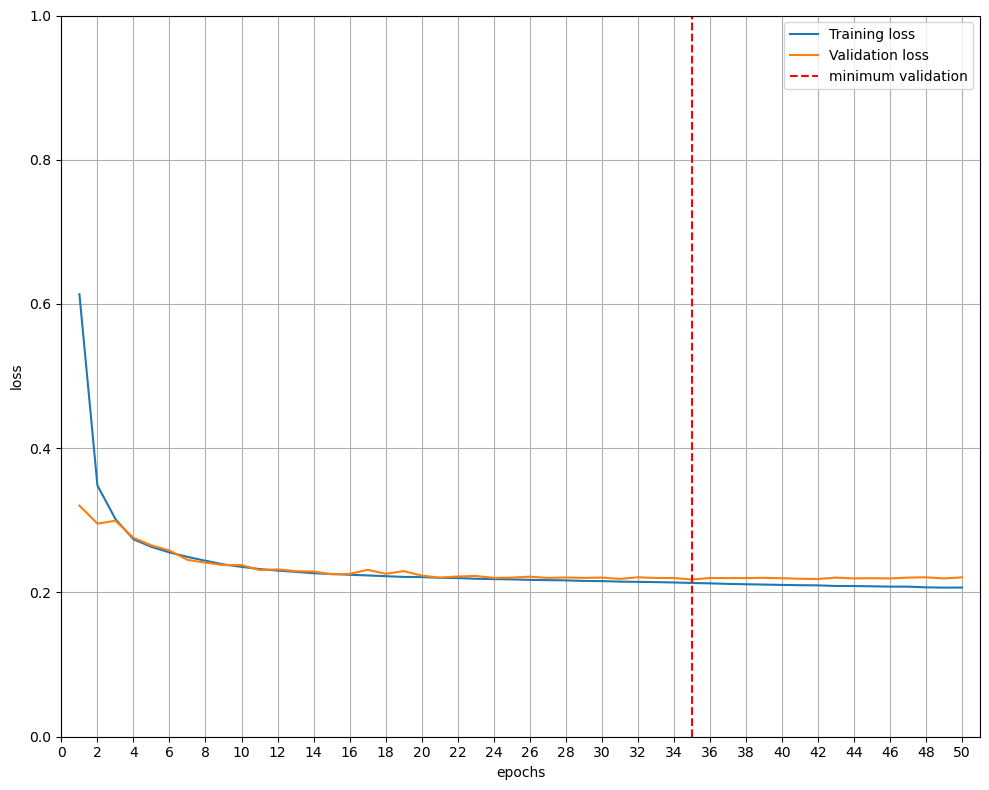}
\caption{Learning curve of U-net model.}
\label{fig:learning_curve}
\end{figure}

The models are trained on $n_{past} = 6$ radar images each, sampled every 15 minutes, with the aim to forecast the next $n_{future} = 6$ images, i.e. up to 1 1/2 hours in the future. The models are implemented in Pytorch. Initial training was done using one NVIDIA T4 GPU on the interactive Google Colab platform, production runs were carried out using one NVIDIA Tesla K80 GPU. Training was done using the Adam optimizer with a learning rate of $lr = 0.0002$. In Figure~\ref{fig:learning_curve}, we show learning curve of U-net model.

For evaluation, we compute (mae) and (f1 score) over the test data for each of the $6$ frames for every prediction. The (mae) metric is defined as the absolute error between the predicted precipitation and the ground truth. The (f1 score) is calculated from the precision and recall and is commonly used by machine learning researchers. To compute (f1 score), we convert the prediction and ground truth to a 0/1 matrix using thresholds of $0.1mm/h$ and $1mm/h$ precipitation rate.

\subsection{Evaluation of loss function}

\begin{figure}[htb]
\begin{minipage}[b]{0.315\linewidth}
  \centering
  \centerline{\includegraphics[width=\linewidth]{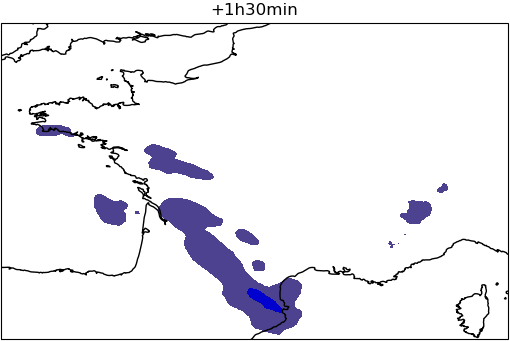}}
  \centerline{\footnotesize (a) mse loss}
\end{minipage}
\begin{minipage}[b]{0.315\linewidth}
  \centering
  \centerline{\includegraphics[width=\linewidth]{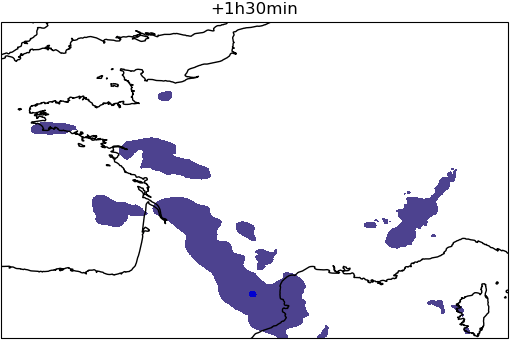}}
  \centerline{\footnotesize (b) wmse loss}
\end{minipage}
\begin{minipage}[b]{0.315\linewidth}
  \centering
  \centerline{\includegraphics[width=\linewidth]{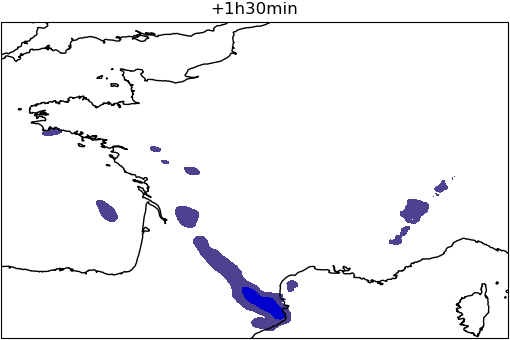}}
  \centerline{\footnotesize (c) ssim loss}
\end{minipage}
\begin{minipage}[b]{0.03\linewidth}
  \centering
  \centerline{\includegraphics[width=\linewidth,keepaspectratio]{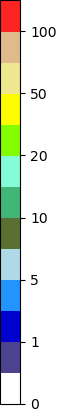}}
  \centerline{}
\end{minipage}
\begin{minipage}[b]{0.315\linewidth}
  \centering
  \centerline{\includegraphics[width=\linewidth]{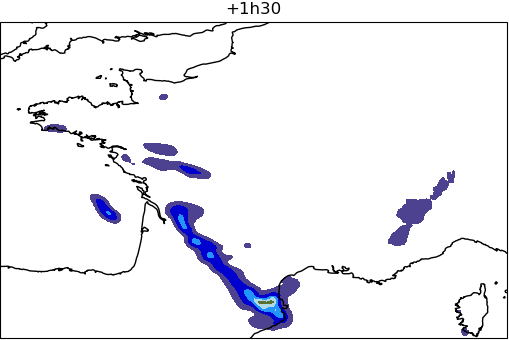}}
  \centerline{\footnotesize (d) wssim loss}
\end{minipage}
\begin{minipage}[b]{0.315\linewidth}
  \centering
  \centerline{\includegraphics[width=\linewidth]{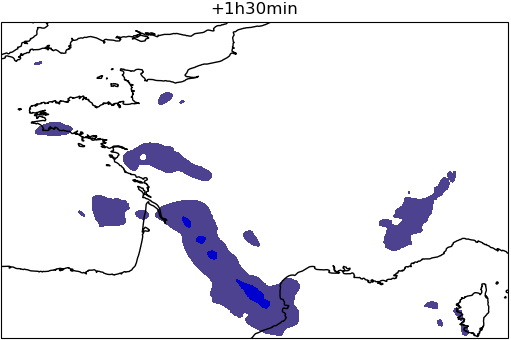}}
  \centerline{\footnotesize (e) wssim+wmse loss}
\end{minipage}
\begin{minipage}[b]{0.315\linewidth}
  \centering
  \centerline{\includegraphics[width=\linewidth]{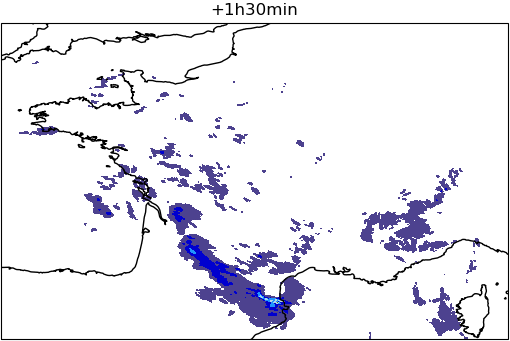}}
  \centerline{\footnotesize (f) GT}
\end{minipage}
\begin{minipage}[b]{0.03\linewidth}
  \centering
  \centerline{\includegraphics[width=\linewidth,keepaspectratio]{images/evaluation_loss/bar.png}}
  \centerline{}
\end{minipage}
\caption{Generated precipitation map using U-net model with different loss functions.}
\label{fig:losses}
\end{figure}

\begin{figure}[htb]
\centering
\begin{minipage}[b]{0.49\linewidth}
  \centering
  \centerline{\includegraphics[width=\linewidth]{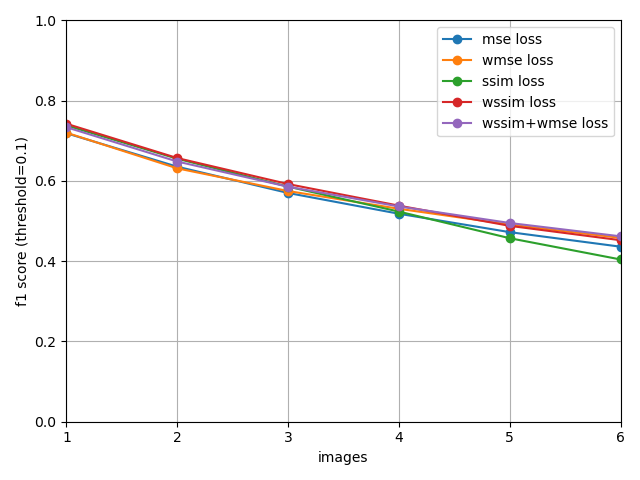}}
    \centerline{\footnotesize (a)}
\end{minipage}
\begin{minipage}[b]{0.49\linewidth}
  \centering
  \centerline{\includegraphics[width=\linewidth]{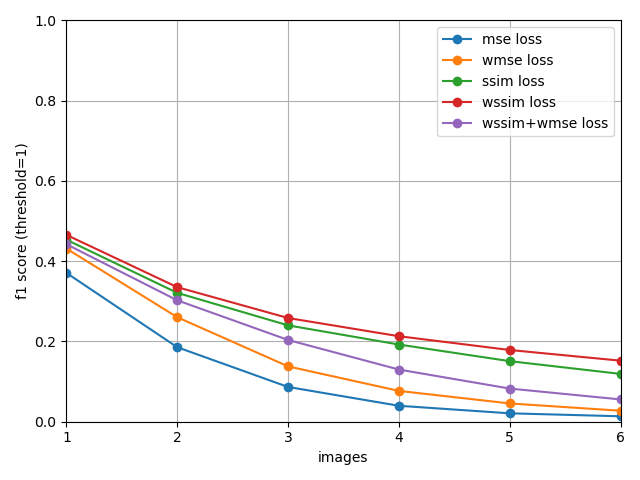}}
 \centerline{\footnotesize (b)}
\end{minipage}
\begin{minipage}[b]{0.49\linewidth}
  \centering
  \centerline{\includegraphics[width=\linewidth]{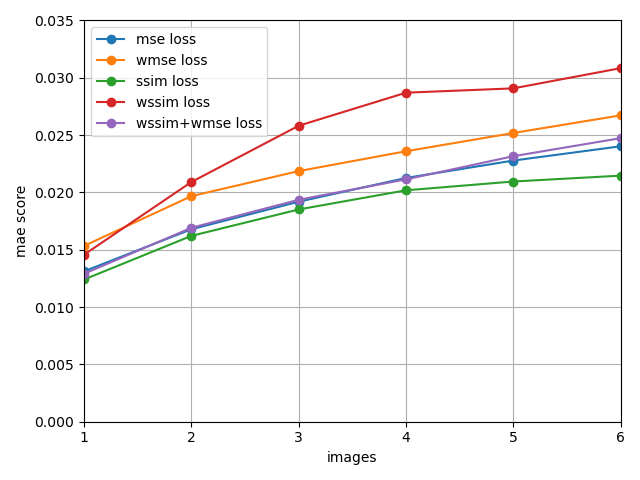}}
    \centerline{\footnotesize (c)}
\end{minipage}
\caption{Quantitative evaluation of loss function using (mae) and (F1-score). The curves shows the scores for 6 outputs frames.}
\label{fig:curves_loss}
\end{figure}

In Figure~\ref{fig:losses}, we show the last output frame (+1h30) of the U-net model using different loss functions. As seen in Figure~\ref{fig:losses}.(b) and Figure~\ref{fig:losses}.(d), the use of weighting in WMSE and WSSIM allows the network to generate more precipitation. Comparing to WMSE, WSSIM loss allows the network to generate highest values of precipitation ($>5mm/h$). as seen in Figure~\ref{fig:losses}.(d), these high values are predicted in the right places comparing to the ground truth image. This can be seen also in (Figure~\ref{fig:curves_loss}.(b)); where the curve of WSSIM loss achieves the highest score of (f1 score) with threshold $1mm/h$. However, WSSIM loss spreads less precipitation in the map comparing to WMSE loss. This can be seen in (Figure \ref{fig:curves_loss}.(c)); where the curve of WSSIM have the highest score of (mae). This justifies the need to combine WSSIM with WMSE losses (Figure~\ref{fig:losses}.(e)).

\subsection{Evaluation of patch extraction method}

\begin{figure}[htb]
\centering
\begin{minipage}[b]{0.49\linewidth}
  \centering
  \centerline{\includegraphics[width=\linewidth]{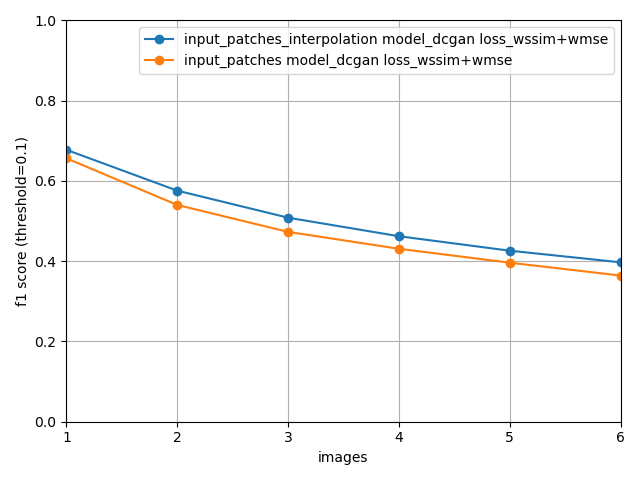}}
  \centerline{\footnotesize (a)}
\end{minipage}
\begin{minipage}[b]{0.49\linewidth}
  \centering
  \centerline{\includegraphics[width=\linewidth]{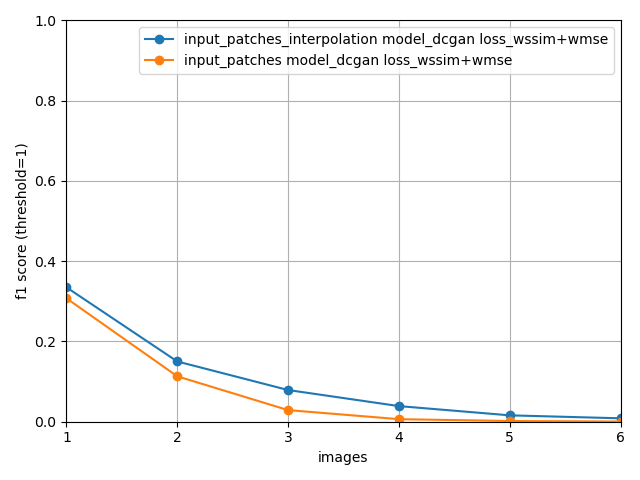}}
\centerline{\footnotesize (b)}
\end{minipage}
\begin{minipage}[b]{0.49\linewidth}
  \centering
  \centerline{\includegraphics[width=\linewidth]{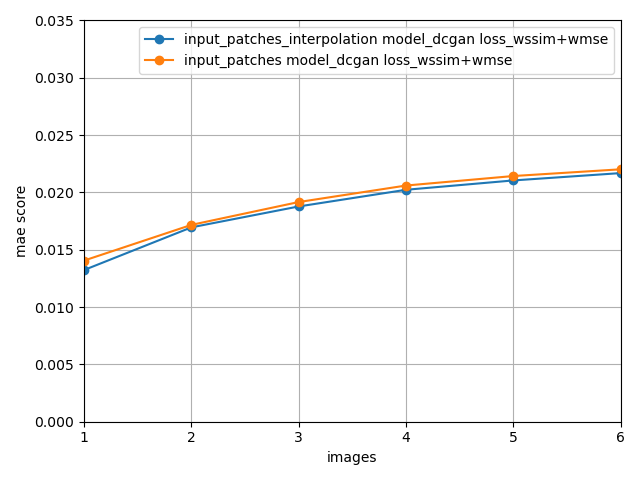}}
  \centerline{\footnotesize (c)}
\end{minipage}
\caption{Quantitative evaluation patch extraction method using (mae) and (F1-score). The curves shows the scores for 6 outputs frames.}
\label{fig:curves_patches}
\end{figure}

For the patch extraction algorithm, we fixed $iSize = 256$, $tSize = 128$, $freq = 20$ and $step = 1$. The size of the interpolated neighboring area is $200$ pixels around the target patch. To study the impact of adding neighborhood information to the patch, we compare two U-net models that take patches as inputs. In the first model, full images are divided into patches of size $256\times256$. Since the target patch size is $128\times128$, then the size of neighbouring region is equal to the margin which is equal to $64$ pixels. In the second model, we use patch extraction method to get patches of $256\times256$. In this model, the margin is also equal to $64$ pixels. However, it summarizes neighboring information of an area of $200$ pixels around the target patch (Figure~\ref{fig:patch_extaction}). As seen in Figure~\ref{fig:curves_patches}, the second model achieves best scores of (mae) and (f1 score).

\subsection{Evaluation of models}

\begin{figure}[htb]
\centering
\begin{minipage}[b]{0.49\linewidth}
  \centering
  \centerline{\includegraphics[width=\linewidth]{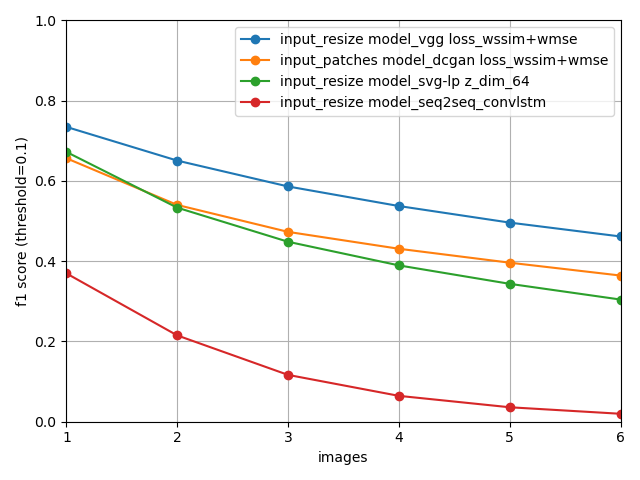}}
  \centerline{\footnotesize (a)}
\end{minipage}
\begin{minipage}[b]{0.49\linewidth}
  \centering
  \centerline{\includegraphics[width=\linewidth]{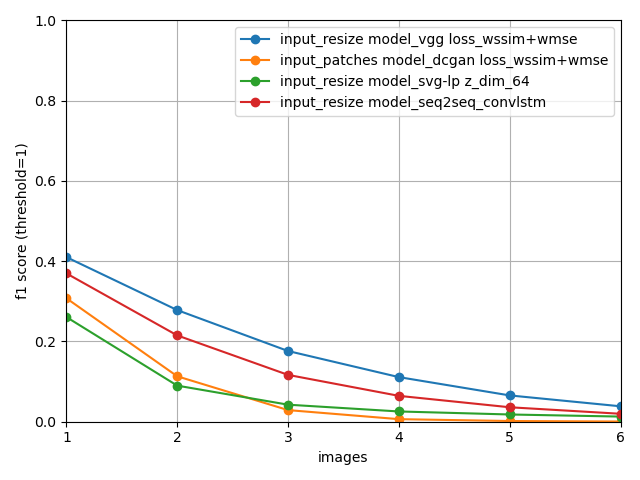}}
  \centerline{\footnotesize (b)}
\end{minipage}
\begin{minipage}[b]{0.49\linewidth}
  \centering
  \centerline{\includegraphics[width=\linewidth]{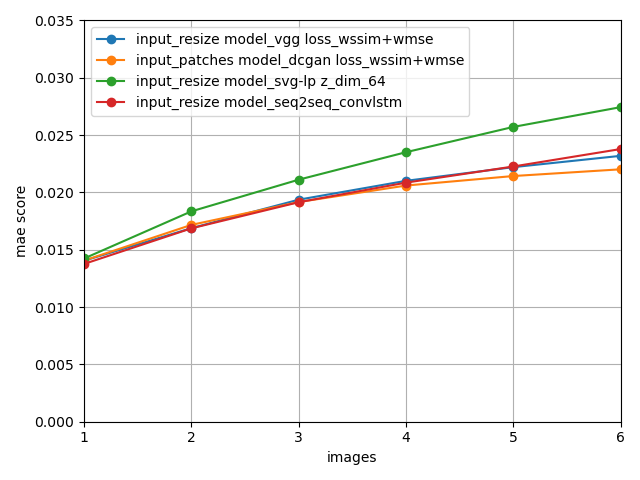}}
  \centerline{\footnotesize (c)}
\end{minipage}
\caption{Quantitative evaluation of the proposed models using (mae) and (F1-score). The curves shows the scores for 6 outputs frames.}
\label{fig:curves}
\end{figure}

\begin{figure*}[htb]
\centering
\begin{minipage}[b]{0.159\linewidth}
  \centering
  \centerline{\includegraphics[width=\textwidth]{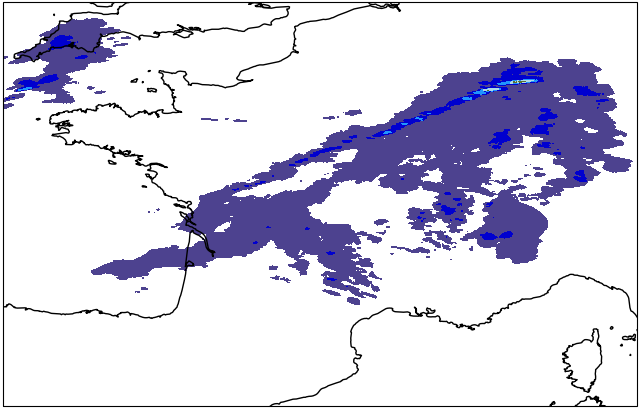}}
  \centerline{\includegraphics[width=\textwidth]{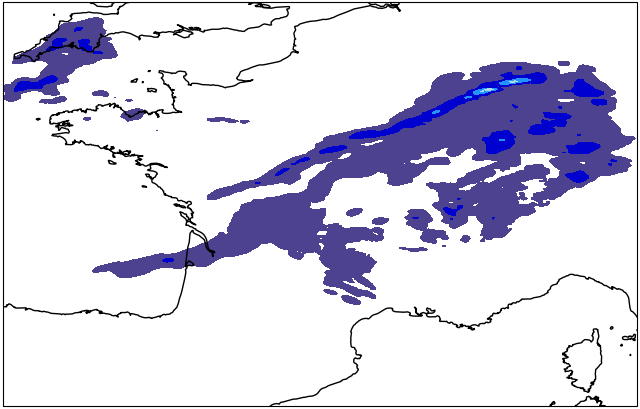}}
  \centerline{\includegraphics[width=\textwidth]{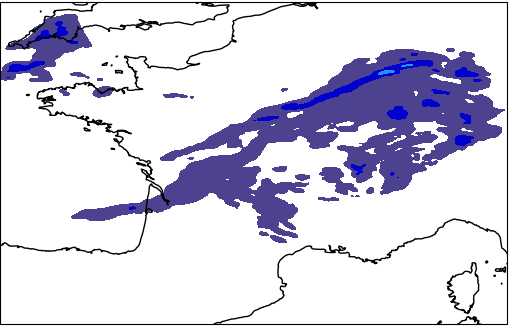}}
  \centerline{\includegraphics[width=\textwidth]{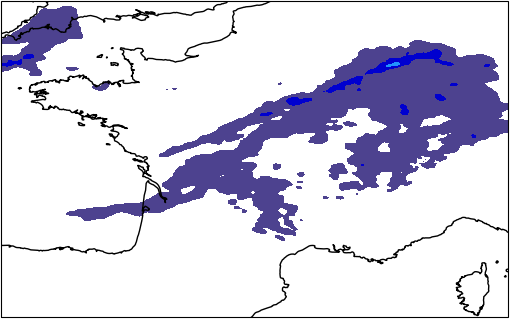}}
  \vspace{0.5cm}
  \centerline{\includegraphics[width=\textwidth]{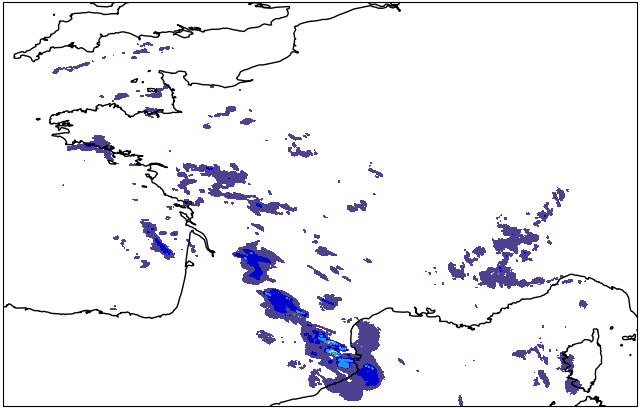}}
  \centerline{\includegraphics[width=\textwidth]{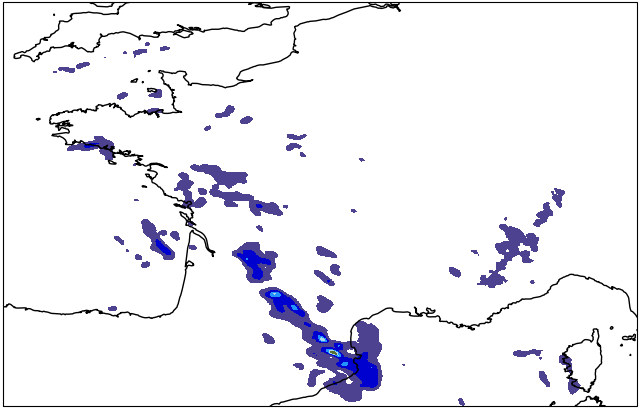}}
  \centerline{\includegraphics[width=\textwidth]{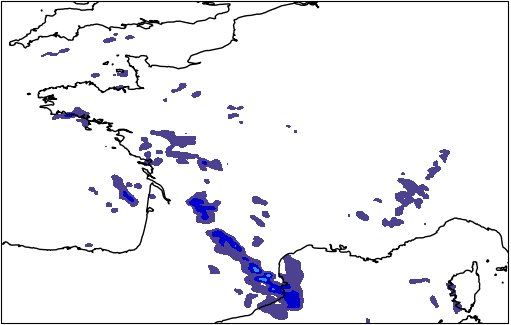}}
  \centerline{\includegraphics[width=\textwidth]{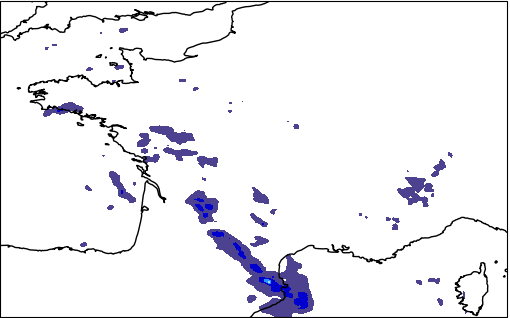}}
  \centerline{\footnotesize +15min}
\end{minipage}
\begin{minipage}[b]{0.1583\linewidth}
  \centering
  \centerline{\includegraphics[width=\textwidth]{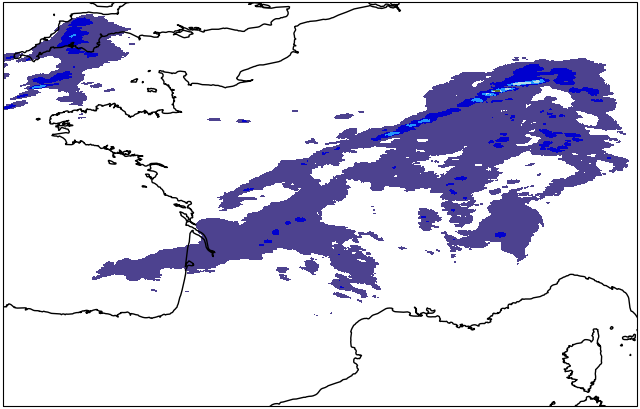}}
  \centerline{\includegraphics[width=\textwidth]{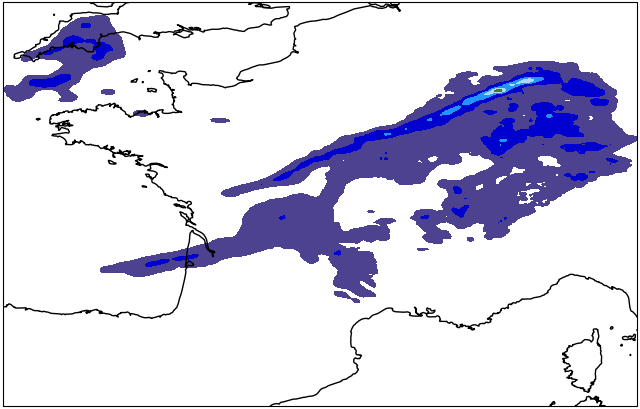}}
  \centerline{\includegraphics[width=\textwidth]{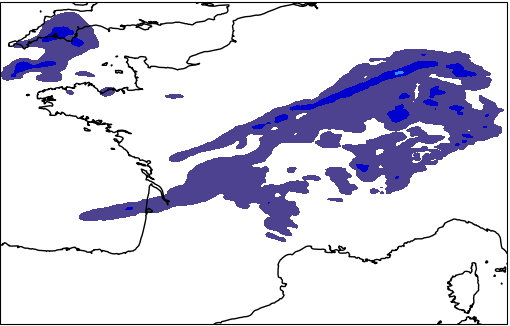}}
  \centerline{\includegraphics[width=\textwidth]{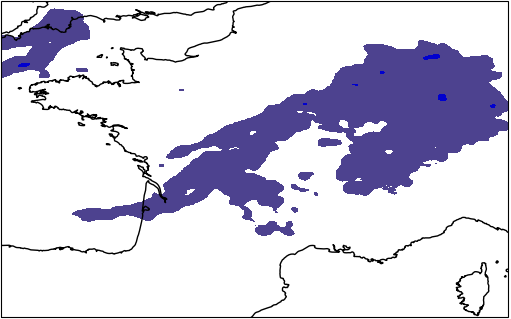}}
  \vspace{0.5cm}
  \centerline{\includegraphics[width=\textwidth]{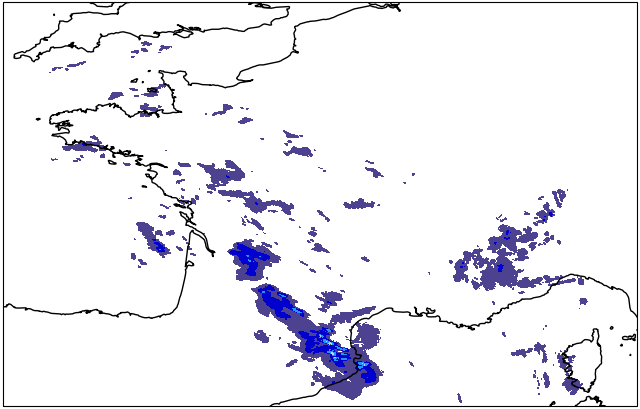}}
  \centerline{\includegraphics[width=\textwidth]{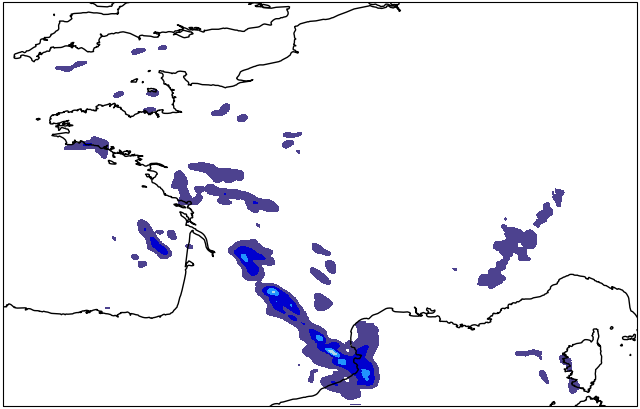}}
  \centerline{\includegraphics[width=\textwidth]{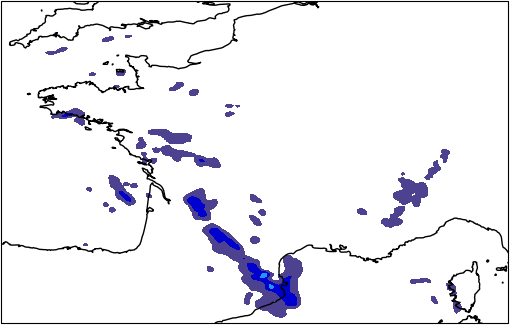}}
  \centerline{\includegraphics[width=\textwidth]{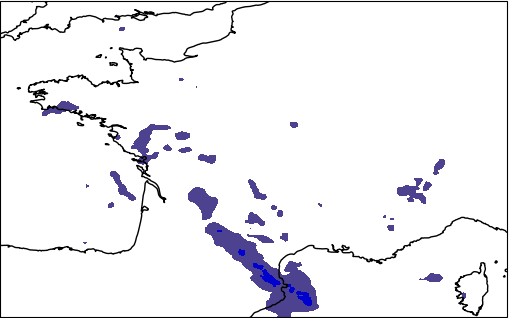}}
  \centerline{\footnotesize +30min}
\end{minipage}
\begin{minipage}[b]{0.1583\linewidth}
  \centering
  \centerline{\includegraphics[width=\textwidth]{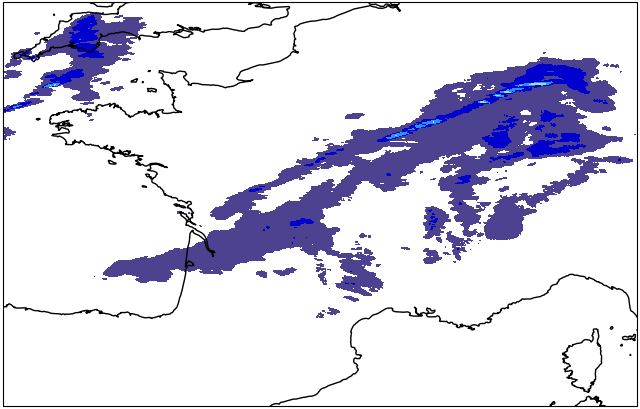}}
  \centerline{\includegraphics[width=\textwidth]{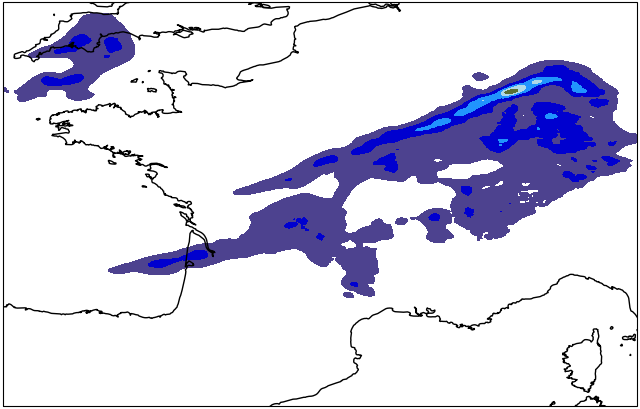}}
  \centerline{\includegraphics[width=\textwidth]{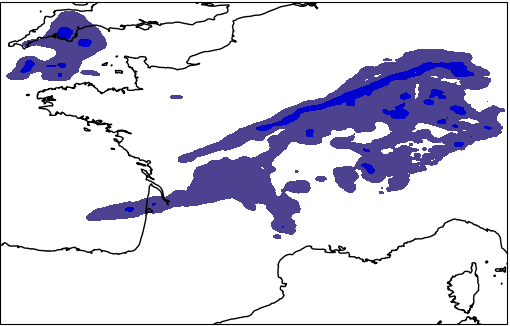}}
  \centerline{\includegraphics[width=\textwidth]{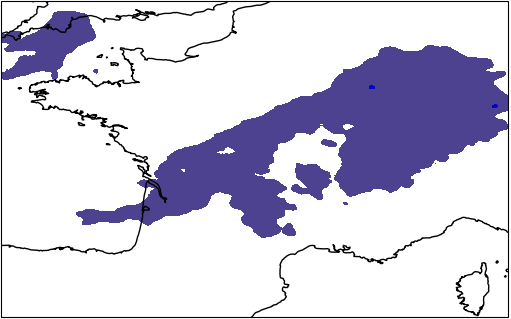}}
 \vspace{0.5cm}
  \centerline{\includegraphics[width=\textwidth]{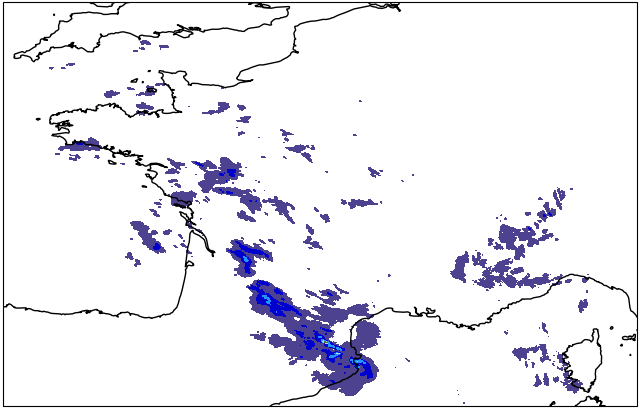}}
  \centerline{\includegraphics[width=\textwidth]{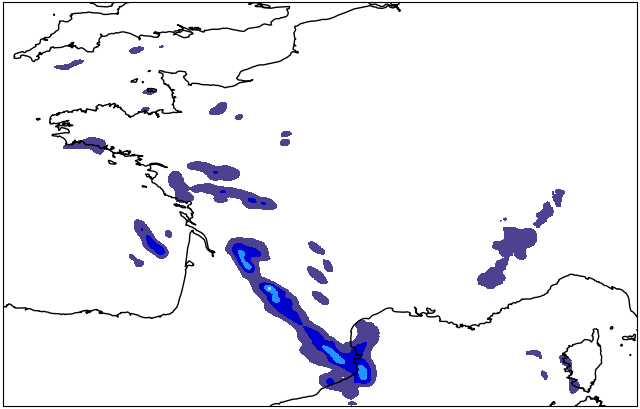}}
  \centerline{\includegraphics[width=\textwidth]{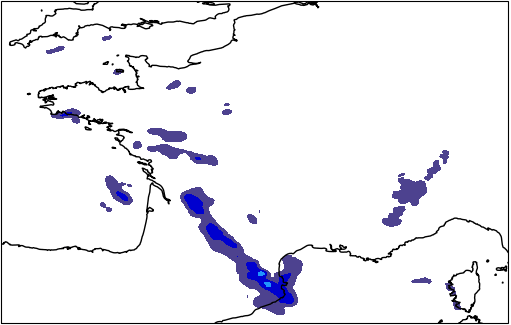}}
  \centerline{\includegraphics[width=\textwidth]{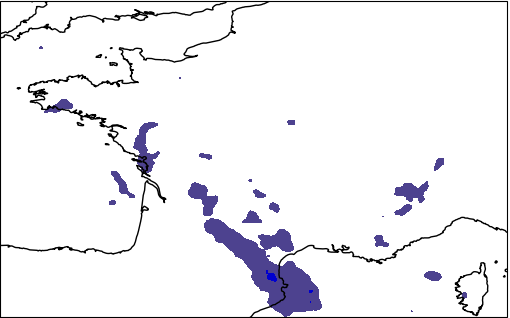}}
  \centerline{\footnotesize +45min}
\end{minipage}
\begin{minipage}[b]{0.1583\linewidth}
  \centering
  \centerline{\includegraphics[width=\textwidth]{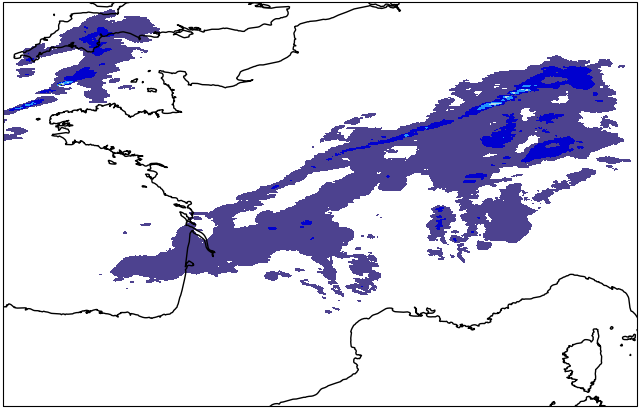}}
  \centerline{\includegraphics[width=\textwidth]{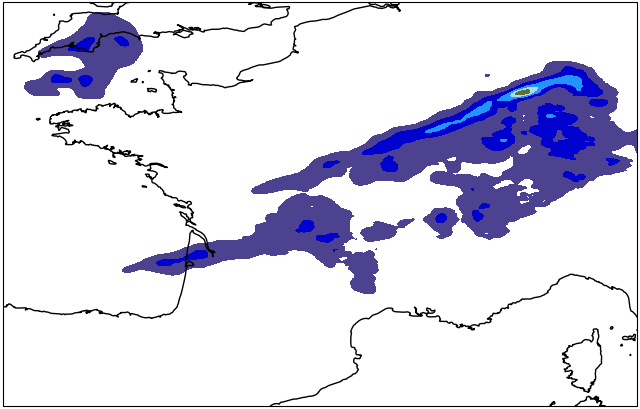}}
  \centerline{\includegraphics[width=\textwidth]{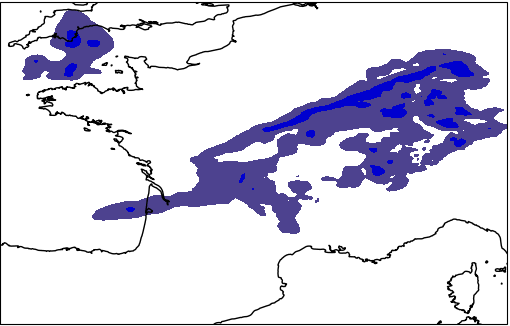}}
  \centerline{\includegraphics[width=\textwidth]{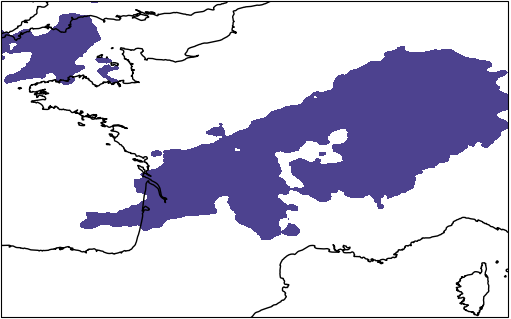}}
 \vspace{0.5cm}
  \centerline{\includegraphics[width=\textwidth]{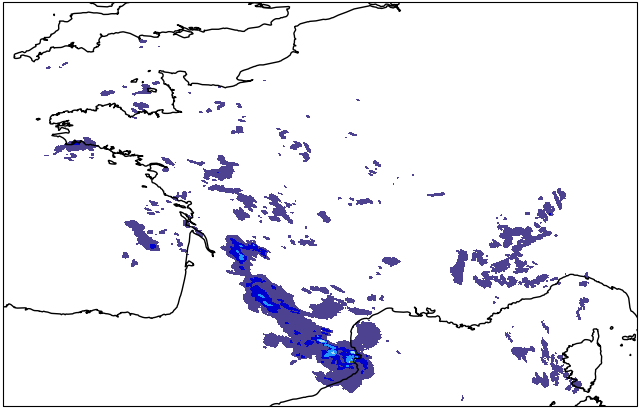}}
  \centerline{\includegraphics[width=\textwidth]{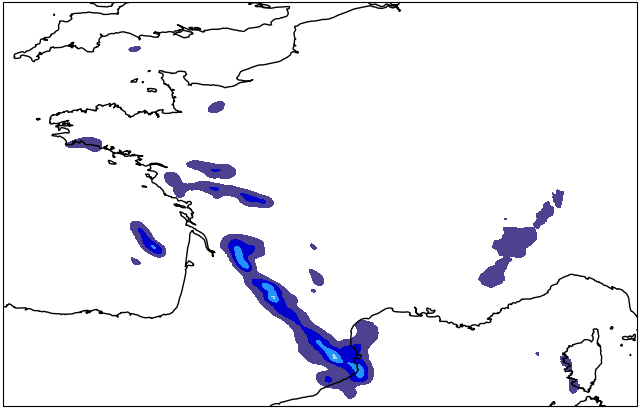}}
  \centerline{\includegraphics[width=\textwidth]{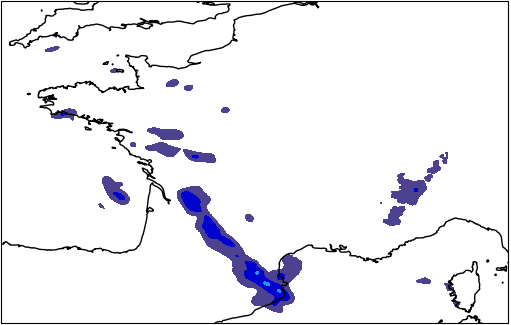}}
  \centerline{\includegraphics[width=\textwidth]{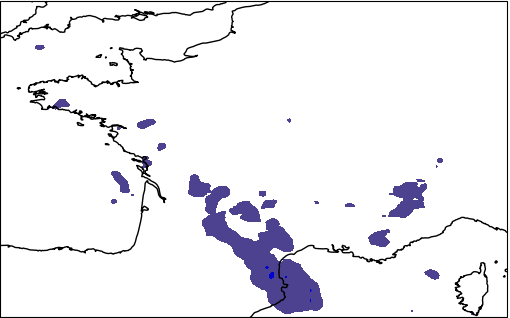}}

  \centerline{\footnotesize +1h}
\end{minipage}
\begin{minipage}[b]{0.1583\linewidth}
  \centering
  \centerline{\includegraphics[width=\textwidth]{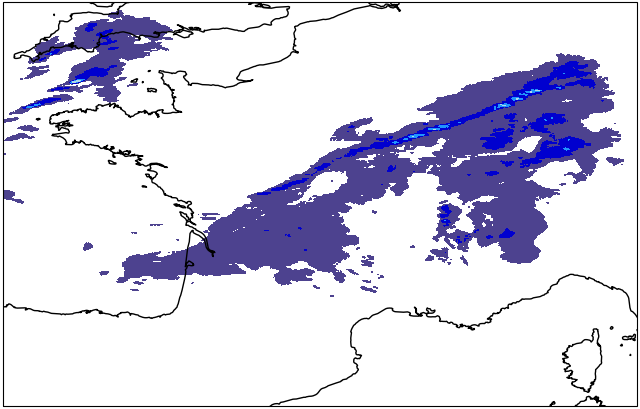}}
  \centerline{\includegraphics[width=\textwidth]{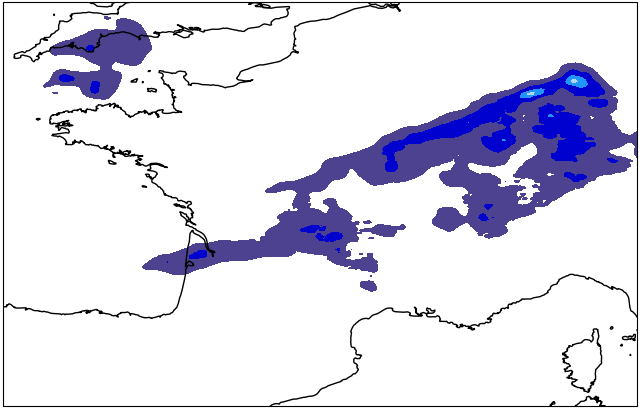}}
  \centerline{\includegraphics[width=\textwidth]{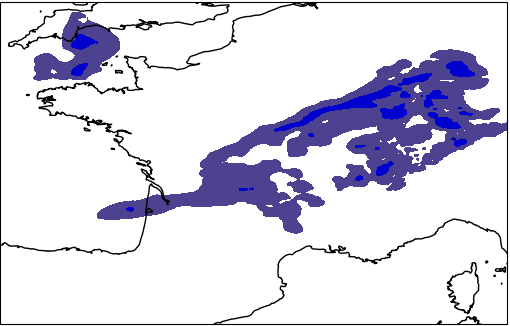}}
  \centerline{\includegraphics[width=\textwidth]{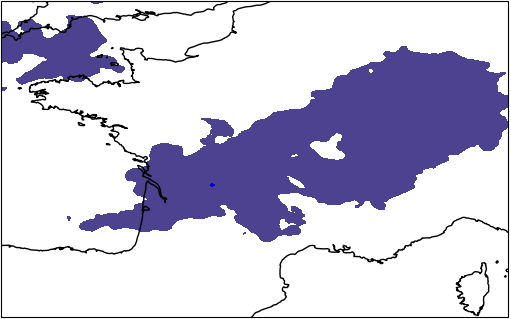}}
  \vspace{0.5cm}
  \centerline{\includegraphics[width=\textwidth]{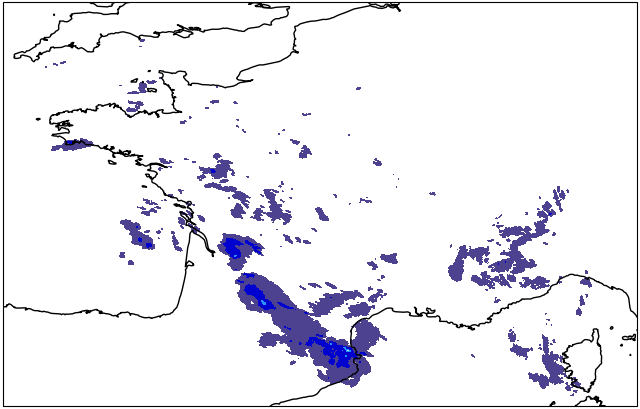}}
  \centerline{\includegraphics[width=\textwidth]{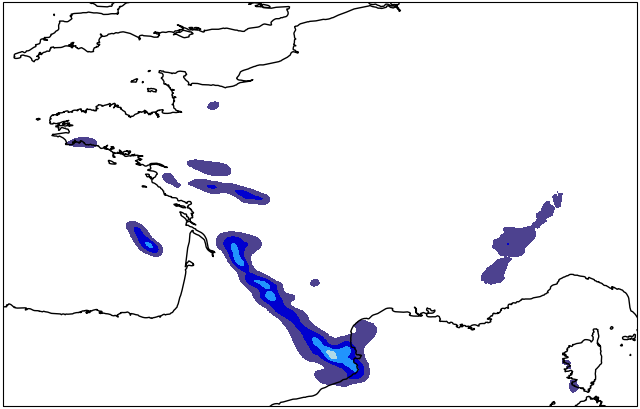}}
  \centerline{\includegraphics[width=\textwidth]{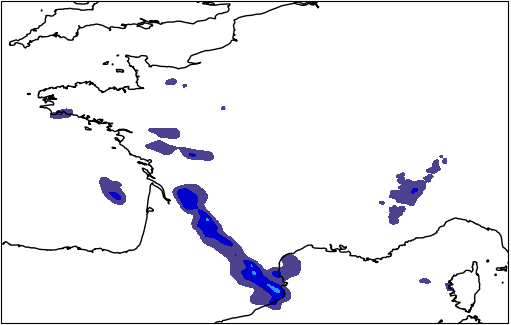}}
    \centerline{\includegraphics[width=\textwidth]{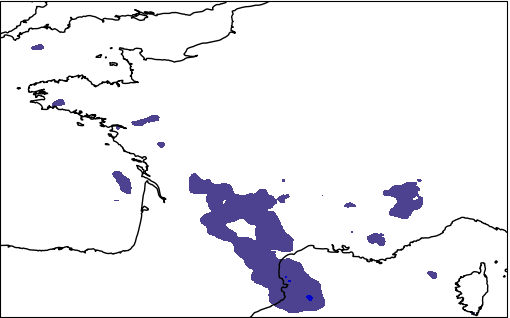}}
  \centerline{\footnotesize +1h15}
\end{minipage}
\begin{minipage}[b]{0.1583\linewidth}
  \centering
  \centerline{\includegraphics[width=\textwidth]{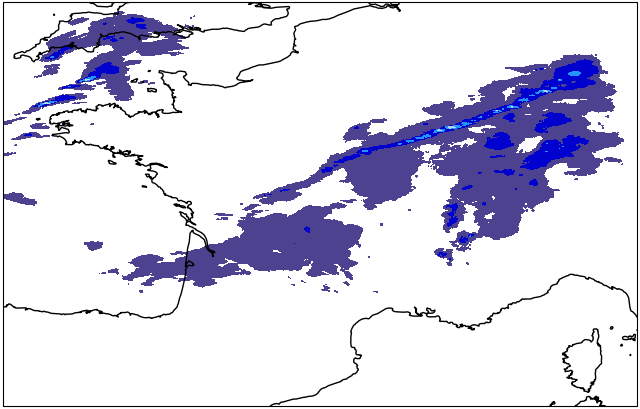}}
  \centerline{\includegraphics[width=\textwidth]{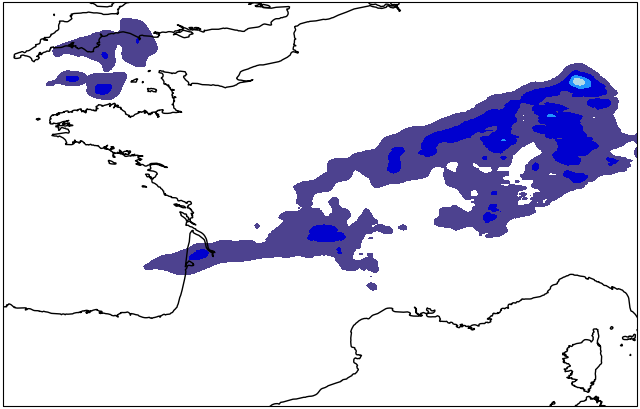}}
  \centerline{\includegraphics[width=\textwidth]{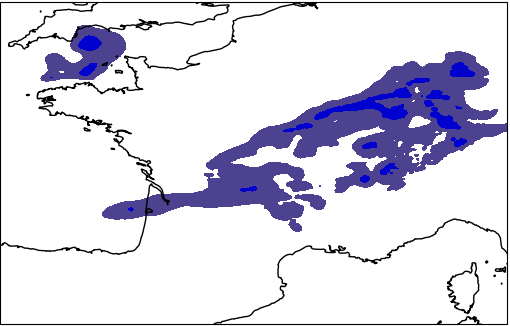}}
  \centerline{\includegraphics[width=\textwidth]{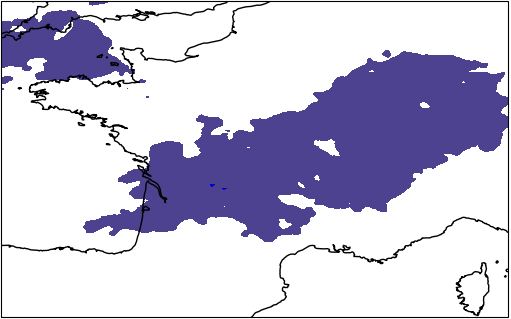}}
  \vspace{0.5cm}
  \centerline{\includegraphics[width=\textwidth]{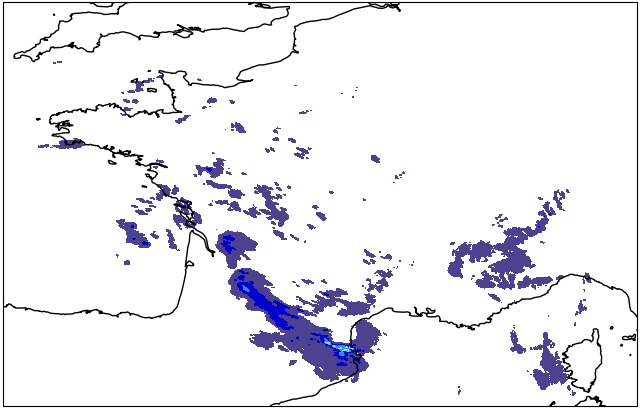}}
  \centerline{\includegraphics[width=\textwidth]{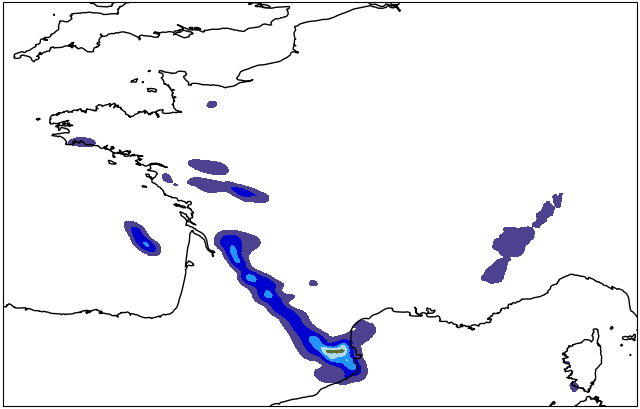}}
  \centerline{\includegraphics[width=\textwidth]{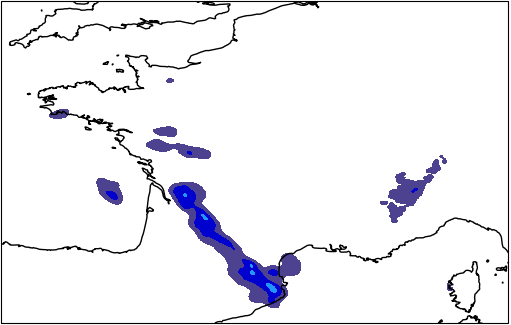}}
  \centerline{\includegraphics[width=\textwidth]{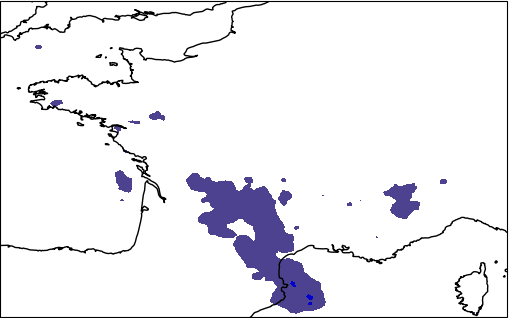}}
  \centerline{\footnotesize +1h30}
\end{minipage}
\begin{minipage}[b]{0.014\linewidth}
  \centering
  \centerline{\includegraphics[width=\linewidth,keepaspectratio]{images/evaluation_loss/bar.png}}
  \centerline{\includegraphics[width=\linewidth,keepaspectratio]{images/evaluation_loss/bar.png}}
  \centerline{\includegraphics[width=\linewidth,keepaspectratio]{images/evaluation_loss/bar.png}}
  \centerline{\includegraphics[width=\linewidth,keepaspectratio]{images/evaluation_loss/bar.png}}
  \vspace{0.5cm}
  \centerline{\includegraphics[width=\linewidth,keepaspectratio]{images/evaluation_loss/bar.png}}
  \centerline{\includegraphics[width=\linewidth,keepaspectratio]{images/evaluation_loss/bar.png}}
  \centerline{\includegraphics[width=\linewidth,keepaspectratio]{images/evaluation_loss/bar.png}}
  \centerline{\includegraphics[width=\linewidth,keepaspectratio]{images/evaluation_loss/bar.png}}
  \centerline{\footnotesize }
\end{minipage}
\caption{Examples of 2 particular precipitation predictions. Top row: Ground truth; second row: U-net model; third row: ConvLSTM model; bottom row: SVG-LP model. The results for the network represent 6 outputs from the model.}
\label{fig:results}
\end{figure*}

In this section, we compare reconstruction quality of the three proposed models: U-net, ConvlSTM and SVG-LP. In Figure~\ref{fig:results}, we present two particular precipitation prediction for the proposed model, which include two types of precipitation. The models predicts 6 image maps in the future. As seen, the three models capture the essential evolution of the precipitation fields. The CNN-based method outperforms the RNN-based models. It is able to generate high value of precipitation. It can predict the future rainfall contour more accurately. It is mainly caused by the strong spatial correlation in the radar maps, i.e., the motion of clouds is highly consistent in a local region. Also, it can be seen that ConvLSTM outperforms SVG-LP. However, ConvLSTM network tends to blur later frames. The blurring effect of ConvLSTM may be caused by the inherent uncertainties of the task. This was one main motivation for the stochastic variational frame prediction method. SVG-LP model triggers more false alarms and is less precise than ConvLSTM. The evaluation scores of the three models are presented in Figure~\ref{fig:curves}. 

\section{Conclusion}
 
In this work, we formulate precipitation nowcasting issue as a video prediction problem where both input and prediction target are image sequences. We trained our models to perform a regression task rather than classification task. Thus, the proposed models generates continuous precipitation values. We evaluated three popular models used in the precipitation nowcasting literature, ConvLSTM, SVG-LP and U-net. We proposed an algorithm to extract patch from precipitation map while including neighboring information for each patch. We proposed a loss function to solve the blurry image issue and to reduce the influence of zero value pixels in precipitation maps. Experiments show that the proposed method captures spatiotemporal correlations and yields meaningful forecasts. In addition, the quantitative evaluation shows promising results.

\bibliography{refs}
\bibliographystyle{icml2019}

\end{document}